\documentclass[review]{elsarticle}

\usepackage{lineno,hyperref}
\usepackage{moreverb}
\usepackage{graphicx, wrapfig}
\usepackage{algorithm}
\usepackage{algpseudocode}
\usepackage{color}
\usepackage{amsmath}
\usepackage{txfonts}
\usepackage{afterpage}
\usepackage[margin=1in]{geometry}
\usepackage{mathrsfs}
\usepackage{tikz}
\usepackage{pstricks}
\usepackage{diagbox}
\usepackage{arydshln}
\usepackage{dsfont}
\usepackage{smartdiagram}
\usepackage{todonotes}
\usepackage{verbatim}
\usepackage{smartdiagram}
\usepackage{bm}
\usepackage{commath}
\usepackage{bbm}
\usepackage{mathrsfs}
\usepackage{dsfont}
\usepackage{extarrows}
\usepackage{subfigure}
\usepackage{multirow}
\usepackage{graphicx}
\usepackage{array}
\usepackage{makecell}
\usepackage{bm}
\usepackage{amsmath} 
\usepackage{commath}
\usepackage{extarrows}
\usepackage{subfigure}
\usepackage{tikz}
\usepackage[toc,page]{appendix}

\newcommand{\bd}{\bm{d}}

\newcommand{\x}{x}
\newcommand{\y}{y}

\newcommand{\s}{s}

\newcommand{\Like}{\bm{L}}

\newcommand{\calL}{\mathcal{L}}
\newcommand{\calD}{\mathcal{D}}
\newcommand{\calB}{\mathcal{B}}
\newcommand{\bdobs}{\bm{d}}
\newcommand{\dobs}{d}

\newcommand{\D}{D}
\newcommand{\Di}{D_i}

\newcommand{\bbR}{\mathbb{R}}

\DeclareMathOperator*{\argmin}{arg\,min}
\DeclareMathOperator*{\argmax}{arg\,max}
\newcommand{\bbE}{\mathbb{E}}
\newcommand{\bbV}{\mathbb{V}}

\newcommand{\obsvar}{\sigma_{\text{obs}}}

\newcommand{\frakN}{\mathfrak{N}}
\newcommand{\hij}{h_{ij}}
\newcommand{\hathij}{\widehat{h}_{ij}}
\newcommand{\G}{G}

\newcommand{\obsoper}{\mathfrak{O}}

\newcommand{\Nx}{N_x}

\newcommand{\Ny}{N_{y}}

\newcommand{\Ndij}{N_{d_{ij}}}
\newcommand{\Ndi}{N_{d_{i}}}
\newcommand{\Lossrec}{\calL^{rec}}
\newcommand{\LossKL}{\calL^{KL}}
\newcommand{\ND}{{N_D}}
\newcommand{\obsnoise}{\epsilon_{\text{obs}}}
\newcommand{\gla}{\alpha} 
\newcommand{\lla}{\beta} 
\newcommand{\lgen}{\mathcal{H}} 
\newcommand{\ggen}{\mathcal{G}} 
\newcommand{\Ngla}{N_{\alpha}} 
\newcommand{\Nlla}{N_{\beta}} 
\newcommand{\Nd}{{N_d}}
\newcommand{\NC}{{N_C}}
\newcommand{\DS}{\mathcal{D}} 
\newcommand{\DSX}{\bm{X}} 
\newcommand{\DSY}{\bm{Y}} 
\newcommand{\TraSetij}{\Lambda_{ij}} 
\newcommand{\TesSetij}{\Delta_{ij}}

\newcommand{\NX}{K} 
\newcommand{\ar}{ar}
\newcommand{\gptol}{\delta_{\text{tol}}}
\newcommand{\kernel}{\mathcal{K}}
\newcommand{\reint}{\epsilon_{ij}^{\text{(int)}}}

\newcommand{\NN}{\mathcal{F}} 

\newcommand{\NKL}{N_{\text{KL}}}

\newcommand{\xblend}{\x_{\text{ble}}}
\newcommand{\xstitch}{\x_{\text{sti}}}
\newcommand{\xtruth}{\x_{\text{truth}}}
\newcommand{\xg}{\x_{\text{g}}}
\newcommand{\gpmean}{m}
\newcommand{\Nbatch}{N_b}

\usetikzlibrary{calc,trees,positioning,arrows,chains,shapes.geometric,%
    decorations.pathreplacing,decorations.pathmorphing,shapes,%
    matrix,shapes.symbols}

\tikzset{
>=stealth',
  punktchain/.style={
    rectangle, 
    rounded corners, 
    draw=black, very thick,
    text width=10em, 
    minimum height=2em, 
    text centered, 
    on chain},
  line/.style={draw, thick, <-},
  element/.style={
    tape,
    top color=white,
    bottom color=blue!50!black!60!,
    minimum width=6em,
    draw=blue!40!black!90, very thick,
    text width=10em, 
    minimum height=3.5em, 
    text centered, 
    on chain},
  every join/.style={->, thick,shorten >=1pt},
  decoration={brace},
  tuborg/.style={decorate},
  tubnode/.style={midway, right=2pt},
}

\tikzstyle{arrow} = [thick,->,>=stealth]

\newdefinition{rmk}{Remark}
\newproof{pf}{Proof}

\newcommand{\comm}[1]{}

\makeatletter
\def\ps@pprintTitle{%
   \let\@oddhead\@empty
   \let\@evenhead\@empty
   \let\@oddfoot\@empty
   \let\@evenfoot\@oddfoot
}
\makeatother

\modulolinenumbers[5]

\journal{Journal of Computational Physics}

\begin{document}

\begin{frontmatter}
		
		\title{A domain-decomposed VAE method for Bayesian inverse problems}

		\author[ShanghaiTechAddress]{Zhihang Xu}
		\ead{xuzhh@shanghaitech.edu.cn}
	\author[AStarAddress]{Yingzhi Xia}
		\ead{Xia_Yingzhi@ihpc.a-star.edu.sg}	
    \author[ShanghaiTechAddress]{Qifeng Liao\corref{mycorrespondingauthor}}
		\cortext[mycorrespondingauthor]{Corresponding author}
		\ead{liaoqf@shanghaitech.edu.cn}
		
		\address[ShanghaiTechAddress]{School of Information Science and Technology, ShanghaiTech University, Shanghai 201210, China}
  \address[AStarAddress]{Institute of High Performance Computing, Agency for Science, Technology and Research (A*STAR), Singapore 138632, Singapore}
\begin{abstract}
Bayesian inverse problems are often computationally challenging when the forward model is governed by complex partial differential equations (PDEs). This is typically caused by expensive forward model evaluations and high-dimensional parameterization of priors. This paper proposes a domain-decomposed variational auto-encoder Markov chain Monte Carlo (DD-VAE-MCMC) method to tackle these challenges simultaneously. Through partitioning the global physical domain into small subdomains, the proposed method first constructs local deterministic generative models based on local historical data, which provide efficient local prior representations. 
Gaussian process models with active learning address the domain decomposition interface conditions. 
Then inversions are conducted on each subdomain independently in parallel and in low-dimensional latent parameter spaces. The local inference solutions are post-processed through the Poisson image blending procedure to result in an efficient global inference result. Numerical examples are provided to demonstrate the performance of the proposed method. 
\end{abstract}
		
	\end{frontmatter}

\section{Introduction}


Inverse problems \cite{kaipio2006statistical} exist in many areas of science and engineering, including the seismic inversion \cite{martin2012stochastic}, the heat conduction \cite{wang2005hierachical} and the inverse groundwater modeling \cite{yeh1986review}. 
The forward model, usually defined through partial differential equations (PDEs), describes certain physical phenomena with parameters as inputs. 
Generally, solving the forward model is computationally expensive but well-defined.
In contrast, the related inverse problem which aims at inferring hidden parameters that cannot be directly observed from limited and noisy observations is typically ill-posed: different sets of parameters can result in similar sensor measurements, and there may be no feasible solution to fit the observed data, or minor errors can render unpredictable changes in the forward model.
The Bayesian methods \cite{tarantola2005inverse,stuart2010inverse},  by viewing the unknown parameters as random variables, formulate the inverse problem into a probabilistic problem to capture the uncertainty in observations, forward models, and prior knowledge.  
One can assign a prior distribution to reflect our knowledge of the parameters before any measurements are made.
The likelihood function is characterized through the forward model.
After collecting observations, Bayes' rule updates the posterior distribution. 
The solution to the Bayesian inverse problem, the posterior distribution, is not a single value but a distribution that can provide statistical information about the inferred parameters. 

The Bayesian perspective provides a framework to characterize the posterior uncertainty, 
but several significant bottlenecks exist when applying the Bayesian formulation in practice.
First, the posterior distribution is usually not available in a closed form but is only known up to a normalization constant. It is then typically approximated via sampling algorithms such as the Markov chain Monte Carlo (MCMC) method \cite{metropolis1953equation, hastings1970monte,robert2013monte}.
Second, when the forward model is described through PDEs, sampling methods that require repeated evaluations of the forward model can cause a severe computational burden. 
Last, the unknowns, which are usually a function or a field, can be high-dimensional after discretization; in some cases, the prior information is only available in the form of historical data. Therefore, proper parameterization is needed to represent the spatially correlated field through relatively low-dimensional random variables. 


In summary, an ideal choice of the parameterization method should: (1) provide dimension reduction for the unknowns; (2) represent the unknowns with the historical data only.
The goal of deep generative models exactly matches this task: generative models can approximate the unknown distribution using its samples with probabilistic models, and deep generative models (DGMs) \cite{salakhutdinov2015learning} formulate the probabilistic models with neural networks to expand the learning ability. 
More importantly, the prior information is naturally explored through training data without other assumptions or restrictions.
The combination of DGMs and inverse problems has recently gained a lot of interest. The work
\cite{laloy2017inversion} proposes to use variational autoencoders (VAEs) \cite{kingma2013auto} as an efficient prior sampling method for inversions in complex geological media.
A multiscale Bayesian inference procedure based on VAEs is presented in \cite{xia2021bayesian}. 
For uncertainty quantification in geophysical inversion, 
the VAE is employed with a deep convolution architecture and  the posterior distribution is exploited using deep mixture models via variational inference \cite{tewari2022subsurface}.
The work \cite{lopez2021deep} experiments with both generative adversarial networks (GANs) \cite{goodfellow2020generative} and VAEs, and suggest that VAEs can be a better choice for inversion due to their ability to balance the accuracy of the generated patterns and the feasibility of gradient-based inversion.

Although the idea of DGMs as a parameterization method is appealing, 
practical situations can be challenging. This includes that
the latent parameter space can still be high-dimensional, and the computational cost for each local forward evaluation is expensive. 
By decomposing the spatial domain into a series of smaller subdomains, 
one can transform the original problem into a set of subproblems with smaller domains and scales that can be easily handled.
In this way, the dimension of the latent parameter space defined in each local subdomain is reduced. 
Furthermore, the computational burden associated with each subdomain is drastically alleviated.

This paper presents a new formulation for Bayesian inversions to handle the challenges mentioned above.
Our contributions are as follows.
First, we propose a domain-decomposed VAE (DD-VAE) method to further expand the representation ability of deep generative priors. 
Our results are based on the simple fully-connected neural networks (FCNNs), demonstrating that a smaller subdomain allows the network architecture to be more adaptive and shallow and enables a more efficient hyperparameter tuning process.
Besides, due to the partition of the spatial domain, the training data size is enlarged, and more importantly, the dimension of the local latent variable is reduced compared to that of the global latent variable.
Second, the computational efficiency after domain division is primarily improved.  
As discussed, repeated evaluations of costly forward models cause a severe computation burden to sampling methods. By decomposing the spatial domain, the global forward model is then transformed into several local forward models, where the computational cost is drastically reduced. 
Moreover, the parallel capability endowed with the domain-decomposed strategy further reduces the computational expense. 
Third, we utilize Gaussian process (GP) models with active learning to manage the interface conditions.  
The last one is a new reconstruction scheme for the unknown global field. 
As directly stitching local fields can cause visible seams over the interfaces, we utilize the Poisson blending technique to blend overlapping local fields to give effective  representations of the global field.

The rest of this paper is organized as follows. 
Section \ref{sec:problem_setup} sets up the problem, where the Bayesian inverse problems and vanilla VAEs are introduced. 
Section \ref{sec:method} introduces our domain-decomposed variational auto-encoder Markov chain Monte Carlo (DD-VAE-MCMC) algorithm. 
Section \ref{sec:numerical_examples} provides numerical examples to verify the efficiency of our proposed method. 
Section \ref{sec:conclusion} concludes this paper.

\section{Problem setup}
\label{sec:problem_setup}
This section briefly reviews the general description of Bayesian inverse problems and the governing problem we consider.

\subsection{The Bayesian inverse problem}
\label{sec:bayes_inv}
Let $\D\subset \bbR^\ND$ ($\ND=1,2,3$) denote a spatial domain that is bounded, connected, and with a polygonal boundary $\partial D$, and $s\in \bbR^{\ND}$ denote a spatial variable.
We consider recovering a spatially-varying parameter function $\x(\s)$ from the $\Nd$-dimensional observed data $\bdobs\in \bbR^{\Nd}$.
For practical reasons, we discretize $\x(\s)$ with the standard finite element method \cite{elman14finite} as a finite-dimensional vector $\x \in \bbR^{N_{\x}}$ ($N_{\x}\gg \Nd$).
The observations and the parameters  usually link through a so-called \textit{forward} problem:
\begin{equation}
    \label{eq:forward_model}
    \bdobs = F(\x)+ \obsnoise\,,
\end{equation} 
where $F$ represents the forward model and $\obsnoise$ denotes an additive noise.
The Bayesian formulation \cite{stuart2010inverse} views the unknown parameter $\x$ as a random variable.
A prior distribution $p(\x)$ of $\x$, which implies the knowledge of the parameter before any measurement, is assumed.  
The probability density function (PDF) of the noise $\obsnoise\in\bbR^{\Nd}$ is denoted by $\pi_{\obsnoise}(\obsnoise)$. 
In this work, the noise $\obsnoise$ is assumed to be Gaussian with zero mean and a diagonal covariance matrix  
$\obsvar^2 I_{\Nd}$, i.e., $\pi_{\obsnoise}(\obsnoise) = \mathcal{N}(\bm{0},\obsvar^2 I_{\Nd})$,
where $\obsvar >0$ is the standard deviation, $I_{\Nd}$ is the identity matrix with size $\Nd\times \Nd$ and $\mathcal{N}(\bm{\mu}, \bm{\Sigma})$ denotes the Gaussian distribution with the mean vector $\bm{\mu}$ and the covariance matrix $\bm{\Sigma}$. 
The likelihood function, which evaluates the discrepancy between the model predictions and the observations, can be given through the definition of the noise, i.e., 
\begin{eqnarray}
\Like(\bdobs|\x) = \pi_{\obsnoise}(\bdobs - F(\x))\propto \exp\left( -\frac{1}{2\obsvar^2}\|F(\x)-\bdobs\|_2^2 \right)\,,
\label{eq_likelihood}
\end{eqnarray}
where $\|\cdot\|_2$ denotes the standard Euclidean norm.
Based on Bayes' rule,  
the posterior distribution of $\x$ can be written as
\begin{equation}
\label{eq:bayes}
\pi(\x|\bdobs) = \frac{ \overbrace{\Like(\bdobs|\x)}^{\text{likelihood}}
    \overbrace{p(\x)}^{\text{prior}}
}
{
    \underbrace{
        \pi(\bdobs)}_{\text{evidence}}}
\propto \Like(\bdobs|\x) p(\x)
\,,
\end{equation}
where the evidence $\pi(\bdobs):= \int \Like(\bdobs|\x) p(\x)\dif \x$ in \eqref{eq:bayes} is usually viewed as a normalization constant.

The posterior distribution \eqref{eq:bayes}, usually does not have an analytical form in the case of nonlinear forward models and has to be approximated with sampling methods such as MCMC \cite{robert2013monte}. 
MCMC generates a Markov chain to approximate the target distribution, which is the posterior distribution in this context. 
To guarantee convergence of MCMC, the detailed balance condition should be satisfied.  
For any two consecutive states $\x$ and $\x^{\star}$, the detailed balance condition is defined as 
\begin{equation}
    \pi(\x|\bdobs) h(\x, \x^{\star}) = \pi(\x^{\star}|\bdobs) h(\x^{\star}, \x)\,,
    \label{eq:detailed_balance} 
\end{equation}
where $h$ denotes the transition kernel.
It is defined as 
\[
    h(\x, \x^{\star}) = Q(\x, \x^{\star})\ar(\x,\x^{\star})\,,
\]
where $Q(\x, \x^{\star})$ is a proposal distribution and $\ar(\x,\x^{\star})$ is the corresponding acceptance probability.
Here, $\x^{\star}$ is proposed by the proposal distribution conditioning on the current state $\x$.
To ensure that the detailed balance condition\eqref{eq:detailed_balance} holds for any prior, likelihood and proposal distribution, the acceptance probability can be defined as 
\[
    \ar(\x,\x^{\star}):= 
    \min\left\{
        1, \frac{\pi(\x^{\star}|\bdobs)Q(\x^{\star},\x)}{ \pi(\x|\bdobs) Q(\x,\x^{\star})}
    \right\}
    = \min\left\{
        1, \frac{\Like(\bdobs|\x^{\star})p(\x^{\star})Q(\x^{\star},\x)}{ \Like(\bdobs|\x)p(\x)Q(\x,\x^{\star})}
    \right\}
    \,.
\] 
Within the basic MCMC framework, the standard Metropolis-Hastings (MH) algorithm \cite{metropolis1953equation,hastings1970monte} is one of the most used sampling techniques for its simplicity. 
However, the main drawback of the MH-MCMC method is that the sampling efficiency degenerates rapidly with an increasing dimension of $\x$.
To this end, we consider the preconditioned Crank Nicolson MCMC (pCN-MCMC) method \cite{beskos2008mcmc,cotter2013mcmc} in this work. 
The proposal distribution is defined to fulfill
\begin{equation}
    Q(\x,\x^{\star})p(\x) = Q(\x^{\star},\x)p(\x^{\star})\,.
    \label{eq:pCN_fullfill}
\end{equation}
Then the acceptance probability $\ar$ can be expressed as 
\begin{equation}
    \ar(\x,\x^{\star}) = \min \left\{1, \frac{\Like(\x^{\star}|\bdobs)}{\Like(\x|\bdobs)} \right\}\,.
    \label{eq:pCN_ar}
\end{equation}
The pCN-MCMC method assumes that the prior $p(\x)$ is a Gaussian distribution, here we let $p(\x)= \mathcal{N}(\bm{\mu},\bm{C})$.
For current state $\x$, the pCN-MCMC proposes $\x^{\star}$ with
\begin{equation}
    \x^{\star} = \sqrt{1-\gamma^2} (\x -\bm{\mu}) + \gamma \zeta + \bm{\mu}\,,
    \label{eq:pCN_proposal}
\end{equation}
where $\zeta\sim \mathcal{N}(\bm{\mu},\bm{C})$ and $\gamma$ is the step size which controls the random movement.
It can be seen that the proposal \eqref{eq:pCN_proposal} satisfies \eqref{eq:pCN_fullfill}.

The procedure of the pCN-MCMC is as follows. 
First one can randomly sample an initial state $\x^{(1)}$ from the Gaussian prior distribution.
Then for the $k$-th state $\x^{(k)}$, a candidate state $\x^{\star}$ is drawn according to \eqref{eq:pCN_proposal}.  
The candidate state $\x^{\star}$ is then accepted as the next state with the acceptance probability $\ar$ (see \eqref{eq:pCN_ar}).
A detailed pCN-MCMC algorithm is summarized in Algorithm \ref{alg:MCMC}. 
\begin{algorithm}[!ht]
    \caption{The pCN-MCMC algorithm}
    \label{alg:MCMC}
    \begin{algorithmic}[1]
        \Require{The forward model $F(\x)$, observational data $\bdobs$, and a prior distribution $p(\x) = \mathcal{N}(\bm{\mu}, \bm{C})$.}
        \State Generate an initial state $\x^{(1)}$ from the prior. 
    \For {$ k =1,\ldots,\NC-1$}
    \State With the step size $\gamma$, 
    propose 
    \[
    \x^{\star} = \sqrt{1-\gamma^2} (\x^{(k)} -\bm{\mu}) + \gamma \zeta + \bm{\mu}\,,
    \]
    where $\zeta\sim \mathcal{N}(\bm{\mu},\bm{C})$.
    \State 
    Compute the acceptance rate 
    	\[
            \ar(\x,\x^{\star}) = \min \left\{1, \frac{\Like(\x^{\star}|\bdobs)}{\Like(\x|\bdobs)} \right\}\,.
\]
where the likelihood  $\Like$
defined in \eqref{eq_likelihood} requires the forward model \eqref{eq:forward_model}.
    \State Draw $\nu$ from a uniform distribution $\nu \sim \mathcal{U}[0,1]$.
    \If {$\nu<\ar(\x,\x^{\star})$}
    \State
    Accept the proposal state, i.e., let $\x^{(k+1)} = \x^\star$.
    \Else
    \State Reject the proposal state, i.e., let $\x^{(k+1)}= \x^{(k)}$.
    \EndIf
    \EndFor
    \Ensure Posterior samples $\{\x^{(k)}\}_{k=1}^{\NC}$.
    \end{algorithmic}
\end{algorithm}

\subsection{PDEs with random inputs}
In this work, we consider a PDE-involved forward model, i.e., $F$ encompasses solving a well-defined PDE system.
The image of $\x$ is denoted by $\Gamma$.
The physics of problems considered are governed by a PDE system over the spatial domain $\D$ and boundary conditions on the boundary $\partial \D$, which are stated as: find
$u(s,\x): D\times \Gamma\to \bbR$, such that
\begin{subequations}
    \begin{align}
    &\calL\left(s,\x;u\left(s,\x\right)\right)=f(s) \,, \qquad
\forall \left(s,\x\right) \in D\times \Gamma, \\
    &\calB\left(s,\x;u\left(s,\x\right)\right)=g(s)\,, \qquad
\forall \left(s,\x\right)\in \partial D\times \Gamma,
    \end{align}
    \label{eq:forward_pde}
\end{subequations}
where $\calL$ is a partial differential operator and $\calB$ is a boundary operator, both of which depend on the random input $\x$.
In addition, $f$ is the source function, and $g$ specifies the boundary conditions. 
Letting $\obsoper$ denote an observation operator, e.g., taking solution values at given sensor locations, 
the observational data $\bdobs$ is defined as $\bdobs = \obsoper(u(s,x))$.
The prior distribution of $\x$ is denoted as $p_{\x}(\x)$, and it is abbreviated as $p(\x)$ without causing ambiguity.

\subsection{The deep generative prior models}
\label{sec:gpm}
In the context of the Bayesian inverse problems, the choice of the prior distribution is crucial yet complex. 
For example, the underlying prior distribution may not be characterized using standard distributions or their variants, or the information about the prior distribution is available implicitly in terms of historical data. 

Deep generative models can serve as a flexible and scalable data-driven parameterization method. 
VAEs and GANs \cite{goodfellow2020generative} can represent the high-dimensional parameter with a low-dimensional latent variable.  
On the contrary, some generative models like flow-based models \cite{dinh2016density} and diffusion models \cite{ho2020denoising} represent the field with a latent variable with the same dimension of the field, which means that the dimensionality issue of the inversion remains. 
In this work, our primary focus is VAEs for their efficiency and simplicity in training compared to GANs.

Consider historical data set $\DSX= \{\x^{(1)}, \ldots,\x^{(\NX)}\} =\{\x^{(k)} \}_{k=1}^\NX$, where $\x^{(k)}\in \bbR^{\Nx}$ are samples independently and identically drawn from the prior distribution $p(\x)$, i.e., $\x^{(k)} \overset{\text{i.i.d.}} {\sim} p(\x)$ for $k=1,\ldots,\NX$.
Letting $p_{\theta}(\x)$ be a probabilistic model where $\theta$ denotes its parameter, 
the goal is to represent the prior distribution $p(\x)$ with $p_{\theta^{\star}}(\x)$ where $\theta^{\star}$ denotes the optimized parameter of the model.
A common criterion for probabilistic models is to maximize the log-likelihood, i.e., $\theta^{\star}:=\argmax_{\theta} \log p_{\theta}(\DSX) = \argmax_{\theta} \sum_{k=1}^{\NX}\log p_{\theta}(\x^{(k)}).$
VAEs introduce a low-dimensional latent variable $\gla \in \bbR^{\Ngla}$ ($\Ngla < \Nx$) to extend the representation capability of the probabilistic models. 
Here, latent variables are variables that cannot be directly observed, and we refer to the joint distribution $p_{\theta}(\x,\gla)$, which are parameterized by neural networks, as the deep latent variable models.
The marginal distribution is then given by $p_{\theta}(\x) = \int p_{\theta}(\x,\gla)\dif \gla$.
However, since the marginal likelihood is an integral and does not have an analytical solution or efficient estimator, it is typically intractable, which causes maximizing the log-likelihood infeasible. 

Due to this intractability, we cannot directly optimize the objective function with gradient descent methods.
The framework of VAEs handles this challenge by introducing a deep parametric model $q_{\phi}(\gla|\x)$ to approximate $p_{\theta}(\gla|\x)$, where $\phi$ is the parameter of the model. 
The model $q_{\phi}(\gla|\x)$ is usually referred to as the \textit{encoder}.
For any given $q_{\phi}(\gla|\x)$, we have
\begin{align}
\log p_{\theta}(x) 
& =\bbE_{q_{\phi}(\gla|\x)}[\log p_{\theta}(\x)] = \bbE_{q_{\phi}(\gla|\x)}\left[  \log \left[ \frac{p_{\theta}(\x,\gla)}{p_{\theta}(\gla|\x)} \right]\right] \nonumber \\
&= \bbE_{q_{\phi}(\gla|\x)}\left[  \log \left[ \frac{p_{\theta}(\x,\gla)}{q_{\phi}(\gla|\x)} \frac{q_{\phi}(\gla|\x)}{p_{\theta}(\gla|\x)} \right]\right]\nonumber\\
& = \underbrace{\bbE_{q_{\phi}(\gla|\x)}\left[  \log \left[ \frac{p_{\theta}(\x,\gla)}{q_{\phi}(\gla|\x)}  \right]\right]}_{\calL_{\theta,\phi}(\x) \text{ (ELBO)}} + 
\underbrace{\bbE_{q_{\phi}(\gla|\x)}\left[  \log \left[  \frac{q_{\phi}(\gla|\x)}{p_{\theta}(\gla|\x)} \right]\right]}_{D_{KL}(q_{\phi}(\gla|\x)\|p_{\theta}(\gla|\x))}\label{eq:vae_loss}\,,
\end{align}
where $\bbE(\cdot)$ denotes the expectation operator and $D_{KL}$ denotes the Kullback-Leibler (KL) divergence. 
The second term in \eqref{eq:vae_loss} is non-negative, then the first term is the lower bound on the log-likelihood of the data:
\begin{align}
\calL_{\theta,\phi}(\x) = \log p_{\theta}(\x) - D_{KL}(q_{\phi}(\gla|\x)\|p_{\theta}(\gla|\x)) \le \log p_{\theta}(\x)\,,
\label{eq:elbo_explain}
\end{align}
this term is also called the \textit{evidence lower bound} (ELBO).
From \eqref{eq:elbo_explain}, we can see that maximizing the ELBO $\calL_{\theta,\phi}$ w.r.t. the parameters $\theta,\phi$ is approximately maximizing the marginal likelihood $p_{\theta}(\x)$ and minimizing the KL divergence from the approximation $q_{\phi}(\gla|\x)$ to the true posterior $p_{\theta}(\gla|\x)$. 

From another perspective, rewrite ELBO \eqref{eq:elbo_explain} as 
\begin{align}
    \calL_{\theta,\phi}(x) 
    & = \bbE_{q_{\phi}(\gla|\x)}\left[  \log \left[ \frac{p_{\theta}(\x,\gla)}{q_{\phi}(\gla|\x)}  \right]\right]
    = \bbE_{q_{\phi}(\gla|\x)}\left[  \log \left[ \frac{p_{\theta}(\x|\gla)p(\gla)}{q_{\phi}(\gla|\x)}  \right]\right]
    \nonumber\\
     & = \underbrace{\bbE_{q_{\phi}(\gla|\x)} 
     \left[\log p_{\theta}(\x|\gla)\right]}_{\calL^{rec}(\x)} +\underbrace{\left\{-D_{KL}\left[q_{\phi}(\gla|\x)\|p(\gla) \right]\right\}}_{\calL^{KL}(\x)}\,.
     \label{eq:elbo}
\end{align}
Thus, the ELBO is composed of a \textit{“reconstruction term”} and a \textit{“regularization term”}. 
The former tends to make the encoding-decoding scheme as performant as possible, and the latter tends to regularize the latent space by making the distribution returned by the encoder close to the given prior $p(\gla)$.
In order to maximize \eqref{eq:elbo}, its gradient $\nabla_{\theta,\phi}\calL_{\theta,\phi}(\x)$ are needed.  
However, they are not available in a closed form and can only be approximated. In the case of continuous latent variables, we can use a \textit{reparameterization trick} for computing unbiased estimates of $\nabla_{\theta,\phi}\calL_{\theta,\phi}(\x)$.
Letting $\epsilon$ denote a random variable whose distribution is independent of $\x$ or $\phi$, one can express $\gla\sim q_{\phi}(\gla|\x)$ as a differentiable and invertible transformation of $\epsilon$: 
$
\gla = \G(\epsilon,\phi,\x)\,.
$
In this work, we specify a standard Gaussian distribution to $\epsilon$ and consider a simple factorized Gaussian encoder, i.e., 
\begin{equation}
q_{\phi}(\gla|\x)=\mathcal{N}(\mu_{\phi}(\x),\text{diag}(\sigma_{\phi}^2(\x)))\,, \quad \gla= \mu_{\phi}(\x) + \sigma_{\phi}(\x)\odot \epsilon\,,\quad 
p(\epsilon) = \mathcal{N}(\bm{0}, \bm{I}_{\Ngla})\,,
\label{eq:repre_trick}
\end{equation} 
where $p(\epsilon)$ denotes the PDF of $\epsilon$,
$\mu_{\phi}(\x)$ and $\sigma_{\phi}(\x)$ are computed by the encoder neural networks,
$\text{diag}(\bm{a})$ denotes the diagonal matrix constructed from the vector $\bm{a}$, 
and $\odot$ denotes the element-wise product. 
To summarize, one can first sample $\epsilon$ from $p(\epsilon)$, then compute $\gla = \G(\epsilon,\phi,\x)=\mu_{\phi}(\x) + \sigma_{\phi}(\x)\odot \epsilon$ to generate a sample $\alpha$ from $q_{\phi}(\alpha|\x)$.
We assign a standard Gaussian distribution for $p(\alpha)$, i.e., $p(\alpha) = \mathcal{N}(\bm{0}, \bm{I}_{\Ngla})$.
Also, we set the decoder $p_{\theta}(\x|\gla)$ as a multivariate Gaussian distribution with identity covariance, i.e., $p_{\theta}(\x|\gla) = \mathcal{N}( \ggen_{\theta}(\gla),\bm{I}_{\Nx})$, where $\ggen_{\theta}(\gla)$ is the output of the decoder network.
As a result, we can easily approximate the reconstruction term via the Monte Carlo estimation,
\begin{align*}
    \Lossrec(\x) 
    & \approx \widetilde{\Lossrec}(\x):=\frac{1}{N_{\text{rec}}}\sum_{k=1}^{N_{\text{rec}}}\log p_{\theta}(\x|\gla^{(k)}) = \frac{1}{N_{\text{rec}}}\sum_{k=1}^{N_{\text{rec}}} -\frac{1}{2}\|\ggen_{\theta}(\gla^{(k)}) - x\|_2^2 + \text{constant}\,,
\end{align*}
where $\gla^{(k)}\sim q_{\phi}(\gla|\x),k=1,\ldots, N_{\text{rec}}$, and the last term is a constant which can be ignored during the optimization process. 
Furthermore, by taking $N_{\text{rec}}=1$, we can form a simple Monte Carlo estimator,
\begin{equation}
     \widetilde{\Lossrec}(\x)= -\frac{1}{2}\|\ggen_{\theta}(\gla) - \x\|_2^2 + \text{constant}\propto  - \|\ggen_{\theta}(\gla) - \x\|_2^2
     \,, \quad \epsilon\sim p(\epsilon), \quad \gla = G(\epsilon,\phi,\x)\,,
    \label{eq:rec_loss}
\end{equation}
Since $p(\gla)$ and $q_{\phi}(\gla|\x)$ are both Gaussian, $\LossKL(\x)$ can be computed analytically, 
\begin{equation*}
    \LossKL(\x)= \frac{1}{2}\sum_{l=1}^{\Ngla}1+\log (\sigma_{\phi}(\x))_l^2 - (\mu_{\phi}(\x))_l^2  - (\sigma_{\phi}(\x))_l^2 \,,
    \label{eq:kl_loss}
\end{equation*}
where $(\mu_{\phi}(\x))_l$ and $(\sigma_{\phi}(\x))_l$ are the $l$-th elements of the vector $\mu_{\phi}(\x)$ and $\sigma_{\phi}(\x)$, respectively.
Then the objective ELBO for single data point $\x$ can be estimated with
\begin{equation*}
    \widetilde{\calL}_{\theta,\phi}(x):=\widetilde{\Lossrec}(\x) + \LossKL(\x)\,.
    \label{eq:elbo_MC}
\end{equation*}
Then the gradients $\nabla_{\theta}\widetilde{\calL}_{\theta,\phi}(x)$, $\nabla_{\phi}\widetilde{\calL}_{\theta,\phi}(x)$ can be effortlessly obtained using auto differentiation. 
The ELBO objective for the data set $\DSX$ is the sum of ELBOs of each individual data point, i.e., $\widetilde{\calL}_{\theta,\phi}(\DSX) = \sum_{\x\in \DSX}\widetilde{\calL}_{\theta,\phi}(\x)$. 
To efficiently minimize $-\widetilde{\calL}_{\theta,\phi}(\DSX)$, 
we consider the mini-batch stochastic gradient descent (SGD) method.

\begin{figure}
    \centering
    \includegraphics[width=0.60\textwidth]{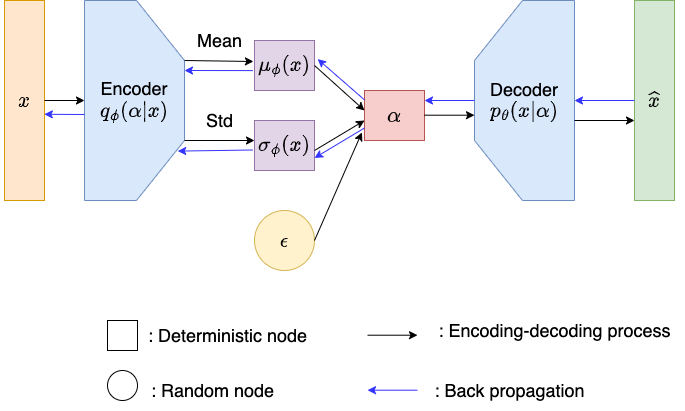}
    \caption{Illustration of the vanilla VAE.}
    \label{fig:vae_ill}
\end{figure}

During the training process, for each epoch, we first randomly draw a batch of data set $\DSX^{b}$ from the complete data set $\DSX$ with batch size $\Nbatch$.
Then for each data point $\x^{(k)}$ in $\DSX^{b}$, we randomly sample noise $\epsilon^{(k)}$ from $p(\epsilon)$. 
We then can compute $\gla^{(k)}$ for $\x^{(k)}$ with the encoder with \eqref{eq:repre_trick}.
The objective for $\DSX^{b}$ is approximated by 
\begin{align}
   -\widetilde{\calL}_{\theta,\phi}(\DSX^b)
    & =
    -\sum_{k=1}^{\Nbatch}\widetilde{\calL}_{\theta,\phi}(\x^{(k)})=
    -\sum_{k=1}^{\Nbatch} \widetilde{\Lossrec}(\x^{(k)}) + \LossKL(\x^{(k)}) \nonumber\\
    & = \sum_{k=1}^{\Nbatch} \left[  \|\ggen_{\theta}(\gla^{(k)}) - \x^{(k)}\|_2^2 -\frac{1}{2}\sum_{l=1}^{\Ngla}1+\log (\sigma_{\phi}(\x^{(k)}))_l^2 - (\mu_{\phi}(\x^{(k)}))_l^2  - (\sigma_{\phi}(\x^{(k)}))_l^2 \right]\,.
    \label{eq:elbo_batch}
\end{align}
Last, we can update $\theta$ and $\phi$ with SGD. 
The training process of the vanilla VAE algorithm \cite{kingma2013auto} is summarized in Algorithm \ref{alg:vae}. 
The illustration for the vanilla VAE is shown in Figure \ref{fig:vae_ill}.

\begin{algorithm}[!ht]
    \caption{The vanilla VAE algorithm}
    \label{alg:vae}
    \begin{algorithmic}[1]
        \Require{The training data set $\DSX$, the encoder $q_{\phi}(\gla|\x)$, the decoder $p_{\theta}(\x|\gla)$, the maximum epoch number $N_{e}$, and the batch size $\Nbatch$.}
        \State Initialize $\theta$ and $\phi$ for the encoder and decoder networks.
        \State Divide $\DSX = \{\x^{(k)} \}_{k=1}^N$ into $n_b$ mini-batches $\{\DSX^{i_b}\}_{i_b=1}^{n_b}$.
    \For{$t=1,\ldots,N_e$}
        \For{$i_b = 1,\ldots,n_b$}
        \State Randomly draw mini-batch training data set $\DSX^{i_b}$ from $\DSX$ with size $\Nbatch$.
        \State Randomly draw noise $\epsilon^{(k)}\sim p(\epsilon)$ for every data point $\x^{(k)}$ in $\DSX^{i_b}$.
        \State Compute the latent variable $\gla^{(k)}$ with the encoder network for every data point $\x^{(k)}$ in $\DSX^{i_b}$ (see \eqref{eq:repre_trick}).
        \State Compute $-\widetilde{\mathcal{L}}_{\theta,\phi}(\DSX^{i_b})$ by \eqref{eq:elbo_batch} and its gradients $\nabla_{\theta}-\widetilde{\calL}_{\theta,\phi}(\DSX^{i_b})$, $\nabla_{\phi}-\widetilde{\calL}_{\theta,\phi}(\DSX^{i_b})$.
        \State Update $\theta$ and $\phi$ using the SGD optimizer.
    \EndFor
    \EndFor
    \Ensure Trained encoder network $q_{\phi^{\star}}(\gla|\x)$ and decoder network $p_{\theta^{\star}}(\x|\gla)$. 
    \end{algorithmic}
\end{algorithm}

\begin{figure}[htbp]
    \centerline{
    \begin{tabular}{c}
    \includegraphics[width=0.65\textwidth]{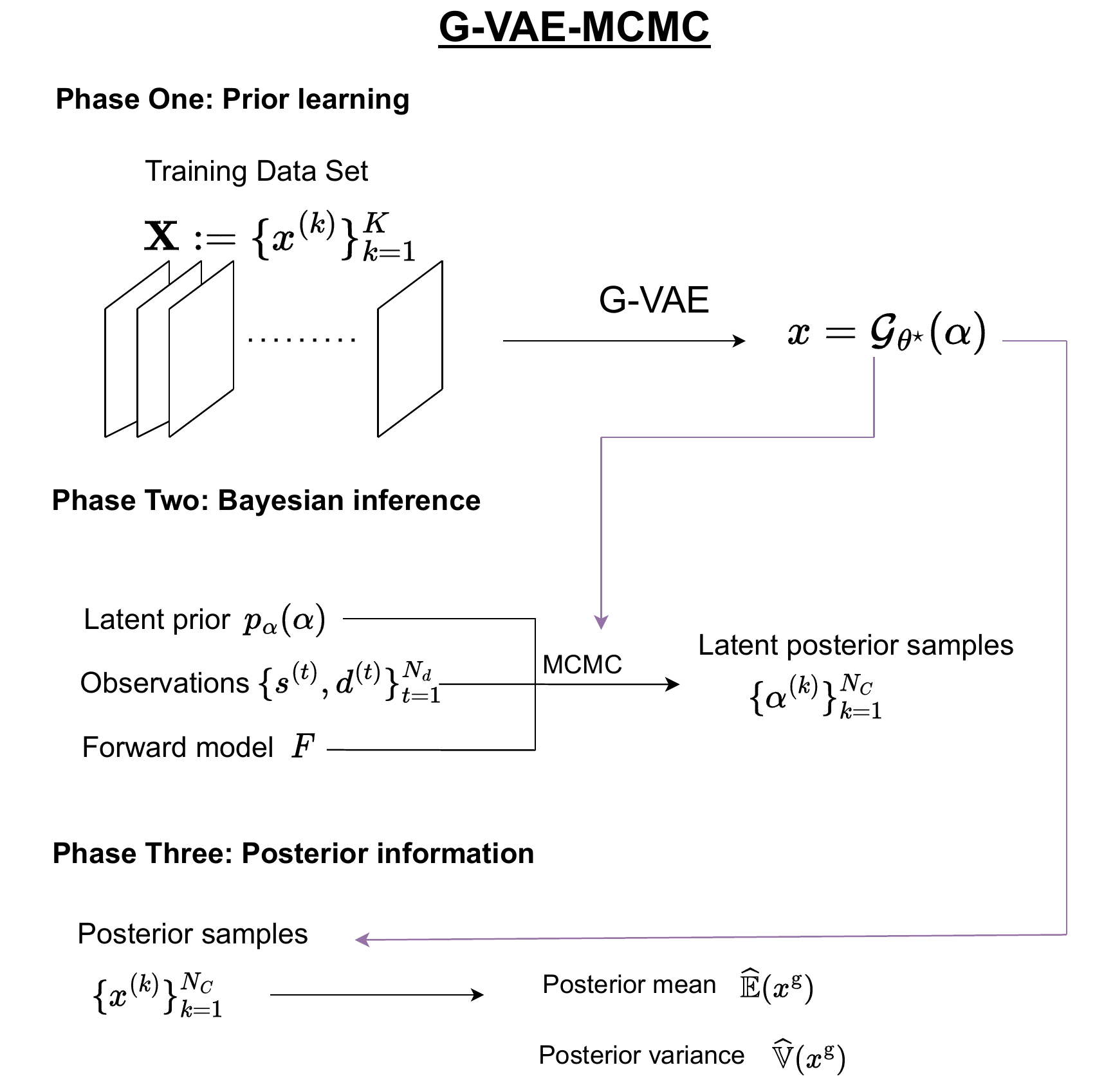}
    \end{tabular}}
    \caption{A diagram of the G-VAE-MCMC method.}
    \label{fig:g-vae-mcmc}
\end{figure}

In this work, we focus on only the mean of the decoder network. 
To sum up, after the training process, to generate samples from the target distribution, one can first sample $\gla\in \bbR^{\Ngla}$ from the given simple low-dimensional probability distribution $p(\gla)$, and then pass it through the trained deterministic generative model to obtain the generative sample $\x$, i.e., 
$
	x = \ggen_{\theta^{\star}}(\gla)\,.
$
Therefore, the inference for $\x$ has turned into the inference for $\gla$,
\begin{equation}
\pi(\gla|\bdobs) \propto \Like(\bdobs|\ggen_{\theta^{\star}}(\gla))p(\gla)\,.
\label{eq:post_z}
\end{equation}
Overall, the inference of high-dimensional parameter $\x$ via deep generative prior roughly has the following steps: 
(a) with training data $\DSX$, one can learn the deterministic generator $\ggen_{\theta^{\star}}$ with Algorithm \ref{alg:vae}; 
(b) given observed data $\bdobs$, the generator $\ggen_{\theta^{\star}}$, the altered forward model $F(\ggen_{\theta^{\star}}(z))$, and a prior distribution $p(\gla)=\mathcal{N}(\bm{0}, \bm{I}_{\Ngla})$, one can use Algorithm \ref{alg:MCMC} to generate posterior latent samples $\{\gla^{(k)} \}_{k=1}^{\NC}$ for the altered posterior distribution \eqref{eq:post_z}; (c) with posterior latent samples, one can generate posterior samples $\{\x^{(k)} \}_{k=1}^{\NC}$ using the generator $\ggen_{\theta^{\star}}$,
and then one can compute posterior statistical information such as the posterior mean and the posterior variance.
For distinction, we refer to the vanilla VAE with training data set defined on the global domain as the \textit{global VAE} method (G-VAE).
Besides, we refer to the global inversion with the G-VAE prior as the global-VAE-MCMC (G-VAE-MCMC) method.
A diagram to illustrate this framework is shown in Figure \ref{fig:g-vae-mcmc}, where the posterior mean and variance are defined in Section \ref{sec:inv_performance}.

\section{A DD-VAE-MCMC method}
\label{sec:method}
The forward system \eqref{eq:forward_pde} can typically be solved by classical numerical methods, e.g., the finite element methods (FEMs) \cite{elman14finite}. However, the corresponding inverse problems remain challenging.  
There are three main bottlenecks when solving such an ill-posed problem. 
The first is the expensive computational cost that occurred in the inversion process. 
The mixing time MCMC takes to ensure convergence is usually very long, and therefore an enormous number of forward computations are required. 
The computational cost can be vast when the forward model is solved by traditional numerical methods. Surrogate models \cite{marzouk2007stochastic,li2014adaptive,liao2019adaptive} are therefore proposed to alleviate this issue at the expense of computational accuracy. 
The second bottleneck is the so-called \textit{curse of dimensionality}. The discretized vector $\x$ is typically high-dimensional and therefore causes extreme sampling difficulties. Parameterization methods, including the Karhunen-Lo\`{e}ve (KL) expansion \cite{spanos1989stochastic}, spring up to handle this problem. 
Even so, the dimensional issue is merely relieved but remains. 
As pointed out in \cite{chen2015local}, the decay rate of eigenvalues depends on the relative length of the spatial domain. In our previous work \cite{xu2022domain}, we propose a local KL expansion method to attack the dimensionality problem.
The third is the difficulty of expressing the prior distribution, as the information about the prior distribution is  not often given in a closed form but is only available through historical samples.

In Section \ref{sec:gpm}, we review the basic settings and the training process of the vanilla VAE.
One can utilize the generative property of VAEs to encode the prior information from given training data.
The deterministic generator defines a mapping from the latent variable space to the complex prior distribution space.
As intriguing as the VAE is, one of the major limitations of such deep models is the enormous computational cost associated with training the neural networks.
In addition, from the perspective of inverse problems, it still has the dimensionality problem, meaning that even if the dimension of the latent variables is drastically reduced, it can be infeasible for MCMC-based sampling methods. 

In this work, we develop a domain-decomposed variational auto-encoder Markov chain Monte Carlo (DD-VAE-MCMC) method to tackle these problems simultaneously. 
The detailed settings of the domain decomposition strategy we adopt are presented in Section \ref{sec:dd_setup}. 
Our domain-decomposed VAE (DD-VAE) method is introduced in Section \ref{sec:DD_VAE}. 
The adaptive Gaussian process method to handle unknown interfaces is discussed in Section \ref{sec:adaptive_gp}.
The image blending technique is reviewed in Section \ref{sec:blending}, and our overall DD-VAE-MCMC algorithm is summarized in \ref{sec:implementation}.

\subsection{The domain decomposition settings}
\label{sec:dd_setup}
\begin{figure}[!htp]
  \centerline{
  \begin{tabular}{c}
  \includegraphics[width=0.50\textwidth]{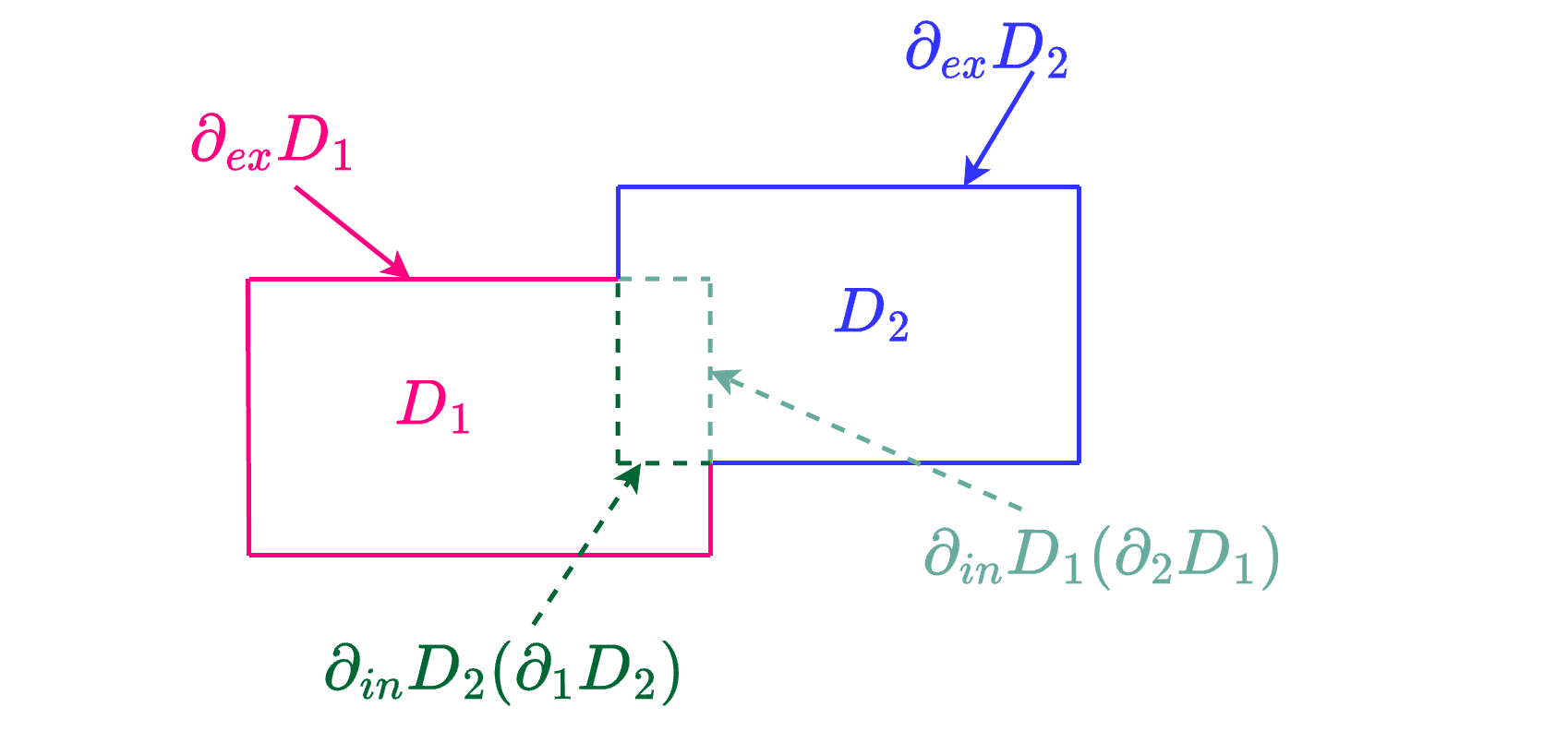}
  \end{tabular}}
  \caption{Illustration of an overlapping two-subdomain system: $\overline{\D} = \overline{\D_1}\cup \overline{\D_2}$, $\partial_{in}D_1=\partial \D_1 \cap \D_2 = \partial_2 \D_1$, $\partial_{in}D_2=\partial \D_2 \cap \D_1 = \partial_1 \D_2$, $\partial_{ex}\D_1 = \partial \D_1 \cap \partial \D$, $\partial_{ex} \D_2 = \partial \D_2 \cap \partial \D$, $\mathfrak{N}_1 = \{2\}$, $\mathfrak{N}_2 = \{ 1\}$ and $\mathfrak{N}= \{(1,2),(2,1)\}$.}
  \label{fig:dd_example}
\end{figure}

Our physical domain $\D \subset \bbR^{N_D}$ is decomposed into $M$ $(M>1)$ overlapping subdomains such that 
\[
  \overline{\D} = \overline{\D_1} \cup \overline{\D_2}\cup \cdots \cup\overline{\D_M}\,,
\]
where the overline  denotes the closure.
For each subdomain $\Di$, ($i=1,\ldots, M$), the set of its boundaries is denoted by $\partial \Di$, and the set of its neighboring subdomain indices is denoted by $\mathfrak{N}_i:=\{j|j\in\{1,\ldots, M\}, j\ne i \text{ and }\Di\cap \D_j\ne \emptyset \}$.
Denoting the interfaces introduced by domain decomposition as $\partial_j \D_i:= \partial \D_i \cap \D_j$, the boundary set $\partial \Di$ can be split into two parts: $\partial \Di = \partial_{ex}\Di \cup \partial_{in}\Di$, where $\partial_{ex}\Di := \partial \Di\cap \partial \D$ are external boundaries and $\partial_{in}\Di:= \cup_{j\in \mathfrak{N}_i}\{\partial_j \D_i \}$ are the interior boundaries.
Grouping all interface indices associated with all subdomains $\{ \Di \}_{i=1}^M$, we define $\frakN:=\{(i,j)|i\in \{1,2,\ldots,M\} \text{ and }j\in \frakN_i \}.$
Figure \ref{fig:dd_example} shows the illustration of notations of an overlapping two-subdomain system.

We introduce decomposed local operators  $\{\calL_i:= \calL|_{\Di}\}_{i=1}^M$, $\{\calB_i:= \calB|_{\partial_{ex}\Di}\}_{i=1}^M$ and local functions $\{f_i:=f|_{\Di}\}_{i=1}^M$, $\{g_i:=g|_{\Di}\}_{i=1}^M$, which are global operators and functions restricted to each subdomain $\Di$.
The restriction of $\x$ to each subdomain is denoted by 
$\x_i:=\x|_{\Di}$.
The dimension of $\x_i$ is denoted as $N_{x_i}$.
In this work, we consider the case where every random variable on each subdomain has the same dimension, i.e., $N_{x_1}=\cdots =N_{x_M}$. 
Suppose that the images of $\{ \x_i\}_{i=1}^M$ are denoted by $\{\Gamma_i\}_{i=1}^M$ respectively and their PDFs are denoted by $p_{\x_i}(\x_i)$, and they are abbreviated as $p(\x_i)$ without ambiguity.
For $(i,j)\in \frakN$, 
$\hij$ denotes the \textit{interface function} defined on the interface $\partial_j\D_i$, and in this work it is defined as $\hij(s,x):=u(s,x)|_{\partial_j\D_i}$, where $u(s,x)$ is the solution of the global problem \eqref{eq:forward_pde}.
Note that the interface function depends on the random input $\x$. 
Then each local problem is defined as: for $i=1,\ldots,M$, find $u_i(\s,\x_i): \Di\times \Gamma_i \to \bbR$ such that
\begin{subequations}
  \begin{align}
    &\calL_i(s,\x_i;u_i(\s,\x_i)) = f_i(\s)\,,\quad \forall (\s,\x_i)\in \Di\times 
    \Gamma_i \,, \\
    &\mathcal{B}_i(\s, \x_i;u_i(\s,\x_i)) = g_i(\s)\,,\quad \forall (\s,\x_i)\in \partial_{ex} \Di\times \Gamma_i\,, \label{eq:sub_problem_boundary}\\
    & \calB_{ij} (\s, \x_i; u_i(\s,\x_i)) = h_{ij}(s,\x)\,,\quad \forall (\s,\x_i)\in \partial_j\D_i \times \Gamma_i, j\in \frakN_i\,,
    \label{eq:sub_problem_interface}
    \end{align}
    \label{eq:sub_problem}
\end{subequations}
where \eqref{eq:sub_problem_interface} defines the boundary conditions on interfaces and $\calB_{ij}$ is an appropriate boundary operator posed on the interface $\partial_j\D_i$.
Detailed discussions for interface functions can be found in \cite{liao2015domain,quarteroni1999domain,chen2015local}. 

\subsection{The DD-VAE method}
\label{sec:DD_VAE}
The idea of combining neural networks with domain decomposition is now actively progressing.
Jagtap et al. proposed extended physics-informed neural networks (XPINNs) \cite{jagtap2020extended} and parallel physics-informed neural networks (parallel PINNs) \cite{jagtap2020conservative} based on domain decomposition. 
As they pointed out, the advantages of the domain decomposition strategy in deep learning-based methods include parallelization capacity, large representation capacity, and efficient hyperparameter tuning. 
\cite{li2019d3m} and \cite{li2023deep} develop deep domain decomposition methods where neural networks are constructed in subdomains with the deep Ritz method. 
Motivated by the  benefits the domain decomposition strategy endows, we in this work propose a domain-decomposed VAE (DD-VAE) method, which embeds the idea of domain decomposition with VAEs.

However, unlike most domain decomposition-based methods in solving PDEs, such as \cite{jagtap2020extended,li2023deep}, the number of neural networks is consistent with the number of subdomains since each neural network needs to contain different physical phenomena. 
In the context of deep generative priors, the training process is not constrained by the boundary conditions and, therefore, can be processed in a \textit{stacking} manner.
For the original training data set $\DSX$, we partition each training data point $\x^{(k)}$ into $\{ \x_i^{(k)}\}_{i=1}^M$ where $\x_i^{(k)}:= \x^{(k)}|_{\D_i}$ in terms of our decomposition strategy of the global domain (see Section \ref{sec:dd_setup}), and in this way, one can obtain the training data set on each subdomain $\DSX_i:=\{ \x_i^{(k)} \}_{k=1}^{\NX}$ for $i=1,\ldots, M.$
We assume that any finite number of the prior distribution follows the same distribution in the case of consistent dimensions.
To be more specific, $\x$ is a collection of random variables with PDF $p(x)$, $\x_i,\x_j$ are two random variables of the same finite number taken from $\x$. 
Then $\x_i$ and $\x_j$ are assumed to follow the same distribution. 
In this work, we consider the case where the dimensions of the local random variables are the same. 
In this way, the trained VAE can be shared over the global domain.
For consistency, the random variable on the subdomain is denoted by $\y$ and the target distribution for local domains as $p(\y)$, i.e., $p_{x_1}(\x_1) = \ldots = p_{x_M}(\x_M)= p_{y}(y)$, where the subscript can be omitted in the absence of ambiguity.
Consequently, all  training data points of subdomains are collected as an augmented training data set $\DSY:= \{ \y^{(k)}| \y^{(k)}\in \DSX_i, i =1,\ldots,M, k=1,\ldots, \NX \}$.


Based on the above assumption, instead of training different generative models with $\DSX_i$, a VAE is trained based on the augmented training data set $\DSY$, which is referred to as the domain-decomposed VAE (DD-VAE). 
$\DSY$ contains $MK$ training data points with dimension $\Ny:=N_{x_1} = \cdots = N_{x_{M}}$. For simplicity, $\DSY$ is rewriten as $\DSY = \{ \y^{(k)}\}_{k=1}^{MK}$.
Following the setup in Section \ref{sec:gpm}, the DD-VAE is developed to learn a local generative model with training data $\DSY$ for the local prior. 
Specifically, the goal here is to learn the distribution $p(\y)$ with probabilistic model $p_{\psi}(\y)$.
We use $\lla\in \bbR^{\Nlla}$ to denote the latent variable in DD-VAE.  
Moreover, the encoder, the decoder, and the deterministic generator are denoted by $q_{\chi}(\lla|\y)$, $p_{\psi}( \y|\lla)$ and $\lgen_{\psi^{\star}}(\lla)$ respectively.

In addition to dimensional reduction, the natural benefits of using our DD-VAE for the prior modeling are: (1) the size of training data is greatly enlarged, (2) for training tasks with smaller domain size, the network can take more superficial structure, (3) through stacking local priors, the overall representation capability is increased. 
To sum up, the training efficiency is greatly improved due to a larger training set and a simpler network structure.

In this paper, fully-connected neural networks (FCNNs) are used. 
Let $\NN:\bbR^{N_i}\to \bbR^{N_o}$ be a feed-forward neural network of $L$ layers,  where $N_i$ and $N_o$ denote the dimension of the input and the output respectively. 
The input variable is denoted by $\eta\in \bbR^{N_i}$ and the $\ell$-th layer is denoted by $\NN_{\ell}: \bbR^{N_{\ell-1}} \to\bbR^{N_\ell}$ ($N_0= N_i$ and $N_L=N_o$).
The weight matrix and the bias vector of the $\ell$-th layer ($1\le \ell \le L$) are denoted by $W_{\ell}$ and $b_{\ell}$, and the overall neural network is a composition of $L$ layers $\NN(\eta) =\NN_L \circ \cdots \circ \NN_1(\eta)$ where 
\[
   \NN_1(\eta) = \Phi(W_1 \eta + b_1)\,,\quad \NN_{\ell} = \Phi(W_{\ell} \NN_{\ell-1}\circ \cdots \NN_1(\eta) + b_{\ell})\,,
\]
where $\Phi(\cdot)$ denotes an activation function.
Collecting all the parameters $\Theta:= \{W_\ell, b_{\ell} \}_{\ell=1}^L$, the parameterized neural network is denoted by  $\NN_{\Theta}$. 
In this work, FCNNs are employed for both encoders and decoders.

\subsection{Adaptive Gaussian process regression interface treatment}
\label{sec:adaptive_gp}
After the spatial domain is decomposed, the interface conditions for local problems \eqref{eq:sub_problem_interface} need to be specified. An adaptive Gaussian process regression method for the interface conditions is developed in our work \cite{xu2022domain}, and we include it here for completeness.  


Denote the set consisting of all the observed data as $\DS = \{ (s^{(t)}, \dobs^{(t)})\}_{t=1}^{\Nd}$, where $s^{(t)}$ is the location of the $t$-th sensor and $\dobs^{(t)}\in \bbR$ is the observation collected at $x^{(t)}$.
The set consisting of all sensor locations is defined as $\bm{s}:=\{s^{(t)}|(s^{(t)},\dobs^{(t)})\in \DS\}$ and the observed data are collected as $\bdobs = [\dobs^{(1)}, \ldots, \dobs^{(\Nd)}]$.
For each subdomain $\Di$ (for $i=1,\ldots, M$), the set consisting of locally observed data is $\DS_i:=\{ (s^{(t)}, \dobs^{(t)})|(s^{(t)}, \dobs^{(t)})\in  \DS \text{ and } s^{(t)}\in D_i \cup \partial D_i \}$, and the size of $\DS_i$ is denoted as $\Ndi:=|\DS_i|$. Similarly, $\bm{s}_i:=\{s^{(t)}|(s^{(t)}, \dobs^{(t)})\in \DS_i\}$ and  
$\bdobs_i\in\bbR^{\Ndi}$ collects the observed data contained in $\DS_i$.

For any $(i,j)\in \frakN$, proper interface functions $\{ h_{ij}(\s,\x)\}_{(i,j)\in \frakN}$ need to be specified.
The interface function virtually depends on the global random parameter $\x$. 
The interface functions for the truth $\x$ are defined as $\hathij(\s) := \hij(\s,\x)$.
Based on limited observational data, an adaptive Gaussian process (GP) \cite{rasmussen2003gaussian} strategy is developed in 
\cite{xu2022domain} to approximate $\hathij$ for its probabilistic formulation.

Suppose that the training data set used to approximate the target function $\hathij(\s)$ is denoted as $\TraSetij = \{(\s^{(t)}, \dobs^{(t)}) \}_{t=1}^{\Ndij}$ with size $\Ndij:= |\TraSetij|$. 
The set consisting of the sensor locations in $\TraSetij$ is denoted by $\bm{\s}_{ij}:= \{s^{(t)}|(s^{(t)},\dobs^{(t)})\in \TraSetij\}$ and $\bd_{ij}\in \bbR^{\Ndij}$ collects all observations in $\TraSetij$.
A Gaussian process is a collection of random variables and any finite combinations of which have a joint Gaussian distribution.
The basic idea is to assume that the target function $\hathij(\s)$ is a realization from a Gaussian random field with mean function $\gpmean(\s)$ and its covariance is specified by a kernel function $\kernel(s,s')$, i.e., the prior GP model is denoted by $\hathij(\s)\sim \mathcal{GP}(\gpmean(\s), \kernel(\s,\s'))$.
Given training data, we want to predict the value of $\hathij(\s)$ at arbitrary point $\s$, which is also Gaussian,
\begin{equation}
  \label{eq:gp_model}
  \hathij(\s)|\TraSetij \sim \mathcal{N}(\gpmean_{\Ndij}(\s), \sigma_{\Ndij}(\s))\,,
\end{equation}
where 
\begin{subequations}
  \begin{align}
  & \gpmean_{\Ndij}(\s) = k_{\star}^T (K_{\Ndij}+\obsvar^2 \bm{I}_{\Ndij})^{-1}\bdobs_{ij}\,,\\
  & \sigma_{\Ndij}(\s) = k(\s,\s) - k_{\star}^T (K_{\Ndij}+\obsvar^2 \bm{I}_{\Ndij})^{-1}k_{\star}\,,
  \label{eq:gp_post_var}
\end{align}
\label{eq:gp_post}
\end{subequations}
where $k_{\star}\in \bbR^{\Ndij}$ and its entries are defined as $(k_{\star})_t = \kernel(\s,\s^{(t)})$ for $\s^{(t)}\in\bm{\s}_{ij}$, $K_{\Ndij}$ is the covariance matrix with entries $[K_{\Ndij}]_{rt} = \kernel(s^{(r)}, s^{(t)})$ for $s^{(r)}, s^{(t)}\in \bm{\s}_{ij}$ and $r,t=1,\ldots,\Ndij$ (see \cite{rasmussen2003gaussian}).

Epistemically, the efficiency of the GP model is greatly dependent on the training data set $\TraSetij$.
Therefore, to maximize the potential efficacy of the GP model, we adaptively construct the GP model. 
First, initial training data set $\TraSetij$ is constructed by randomly choosing one data pair $\{\s^{(t)}, \dobs^{(t)}\}$ from $\DS_i\cup \DS_j$.
In the meantime, a test data set $\TesSetij\subset \partial_j\D_i$ is constructed. 
Then a GP model \eqref{eq:gp_model} is initialized with the current training data set. 
Second, variances of the current GP model are computed for each test point $\s\in \TesSetij$ using \eqref{eq:gp_post_var}, and the test point with the largest variance is denoted as 
\begin{equation}
\overline{\s} := \argmax_{\s \in \Delta_{ij}} \sigma_{\Ndij}(\s).
\end{equation}
Then, we trace the location of the observation which is  the closest to $\overline{\s}$, i.e., 
\[
 \s^{\star}:=\argmin_{\s \in \bm{s}}\|\s - \overline{\s} \|_2\,.
\]
In the third step, the data pair $(\s^{\star}, \dobs^{\star})$ is augmented to the training data set $\TraSetij$, where $\dobs^{\star}$ is the observation collected at $\s^{\star}$.
The GP model is then updated with the current training data set.
The second and third steps are repeated until the maximum posterior variance of the test data set is less than a given threshold $\gptol$. 
The procedure is represented in our main algorithm in Section \ref{sec:implementation}.

With the interface GP models, the local problem \eqref{eq:sub_problem} discussed in Section \ref{sec:dd_setup} is reformulated as:
for $i=1,\ldots,M$, find $u_i^{\text{GP}}(\s,\x_i): \Di\times \Gamma_i \to \bbR$ such that
\begin{subequations}
  \begin{align}
    &\calL_i(s,\x_i;u_i^{\text{GP}}(\s,\x_i)) = f_i(\s)\,,\quad \forall (\s,\x_i)\in \Di\times 
    \Gamma_i \,, \\
    &\mathcal{B}_i(\s, \x_i;u_i^{\text{GP}}(\s,\x_i)) = g_i(\s)\,,\quad \forall (\s,\x_i)\in \partial_{ex} \Di\times \Gamma_i\,, \label{eq:sub_problem_boundary_gp}\\
    & \calB_{ij} (\s, \x_i; u_i^{\text{GP}}(\s,\x_i)) = \gpmean_{\Ndij}(\s)\,,\quad \forall (\s,\x_i)\in \partial_j\D_i \times \Gamma_i, j\in \frakN_i\,,
    \label{eq:sub_problem_interface_gp}
    \end{align}
    \label{eq:sub_problem_gp}
\end{subequations}
where $\gpmean_{\Ndij}(\s)$ is the mean function of GP interface models defined in \eqref{eq:gp_post}. 
Up to now, the local forward model associated with \eqref{eq:sub_problem_gp} is denoted by  $F^{\text{GP}}_i(\x_i):= \obsoper_i(u^{\text{GP}}_i)$ where $\obsoper_i$ is the local observation operator.

\subsection{Image blending}
\label{sec:blending}
Given local samples $\x_i$, $i=1,\ldots,M$, directly stitching the local samples together to obtain a global sample can result in visible seams on the interfaces.
To mitigate this issue, we use image blending techniques which are widely used in computer vision.
In this work, we focus on \textit{the Poisson blending} \cite{perez2003poisson} method (also known as the gradient-domain composition) to blend several samples seamlessly. 
The key idea behind the Poisson blending method is that slow gradients of intensity, which the Laplacian operator can characterize, can be superimposed on an image with barely noticeable effects.
Then \cite{perez2003poisson} formulated the original image editing problem as a Poisson partial differential equation with Dirichlet boundary conditions.
Solving the corresponding PDE gives the blended image.

\begin{figure}[!htp]
  \centerline{
  \begin{tabular}{c}
  \includegraphics[width=0.50\textwidth]{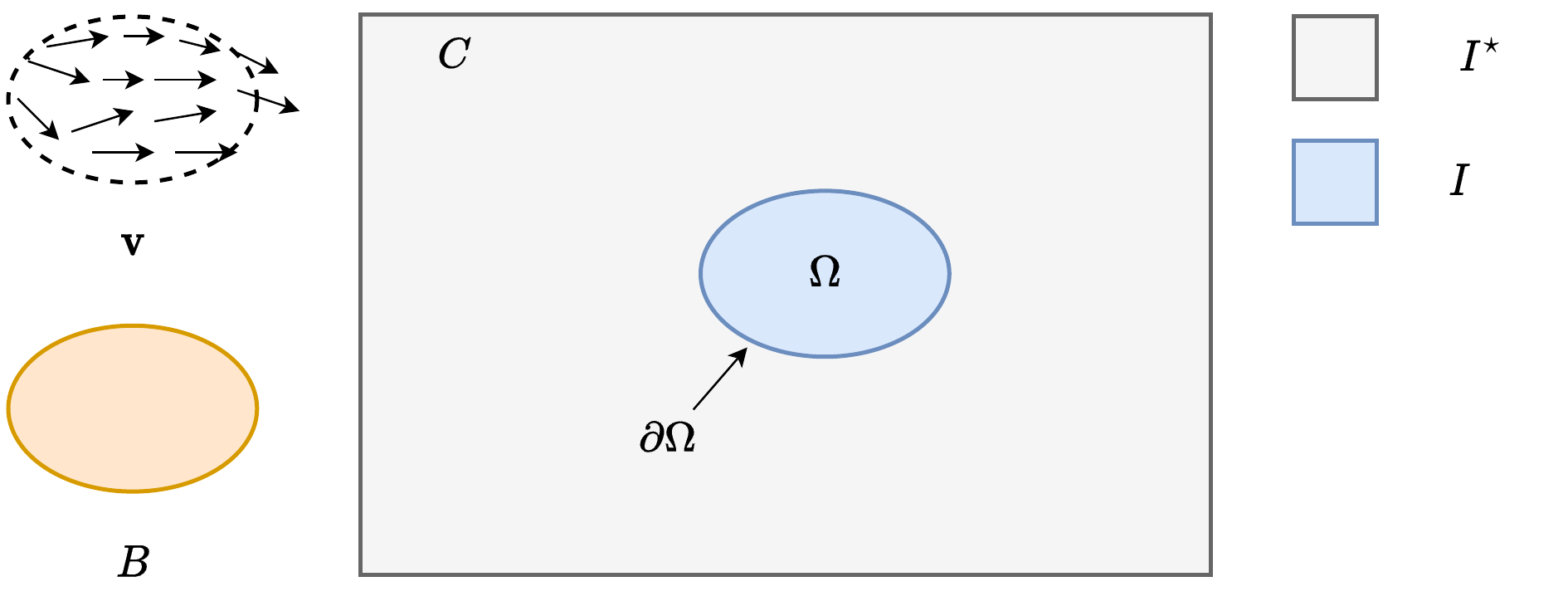}
  \end{tabular}}
  \caption{Illustration of the Poisson image blending method.}
  \label{fig:possion_ill}
\end{figure}

Let $C\subset \bbR^2$ be the image definition domain, and let $\Omega$ be a closed subset of $C$ with boundary $\partial \Omega$.
Let $B$ denote the source image, which is the image we want to insert in $\Omega$, and $\mathbf{v}$ denote its gradient, which is also called the \textit{guidence field}.
Let $I^\star$ denote the original image defined over $C$ minus the interior of $\Omega$ and $I$ denote the unknown scalar function of the generated image defined over the interior of $\Omega$.
Figure \ref{fig:possion_ill} illustrates detailed settings.

The Poisson image editing technique inserts a vector $\mathbf{v}$ to blend pixels in the source image, and so the blending of two images can be obtained through the minimization problem:
\begin{align*}
  &\min_{I} \iint_{\Omega}|\nabla I- \mathbf{v}|^2\,,\quad I|_{\partial \Omega} = I^{\star}|_{\partial \Omega}\,,\\
  & \mathbf{v} = \nabla g\,.
\end{align*}
The solution of the above optimization problem is the unique solution of the following Poisson equation with Dirichlet boundary conditions:
\begin{equation}
\Delta I = \Delta B \text{ over }\Omega\,, \quad{ \text{with} }\quad I|_{\partial \Omega} = I^{\star}|_{\partial \Omega}\,,
\label{eq:poss_min}
\end{equation}
where the boundary condition constrains that the pixel intensity of the generated image should equal the boundary pixel of the target image. 
By minimizing the gradient of the blended image, the source image $B$ is cloned into the target image with a natural appearance. 

We can then represent \eqref{eq:poss_min} in a discrete form.
For each pixel $\s$ in $C$, let $N_\s$ be the set of its 4-connected neighbors which are in $C$, and let $\langle \s,\s'\rangle$ denote a pixel pair such that $\s'\in N_\s$.
The boundary of $\Omega$ is then denoted as $\partial \Omega=\{\s\in C/\Omega: N_s\cap \Omega \ne \emptyset \}$.
Let $I(\s)$ denote the intensity of $I$ at pixel $\s$.
The discrete form of \eqref{eq:poss_min} is 
\[
  \min_{I|_{\Omega}} \sum_{\langle \s,\s' \rangle\cap \Omega\ne 0} (I(\s) - I(\s') -v_{s,s'})^2\,,\quad \text{with }I(\s) = I^{\star}(\s) \text{ for all }\s\in \partial \Omega\,,
 \]
where $v_{s,s'}$ is the first-order derivative of the source image.
Thus, the intensity of each pixel in $\Omega$ in the generated image is determined by the following equations,
\[
  |N_\s|I(\s) - \sum_{s'\in N_s\cap \Omega}I(s') = \sum_{s'\in N_s\cap\partial \Omega}I^{\star}(s') + \sum_{s'\in N_s} v_{s,s'}\,,\quad \forall \s\in \Omega\,.
\]
Solving the discrete Poisson equation results in generated seamless images.
As we can see, the Poisson image blending technique is actually a flexible framework and enables different ways to implement blending with different settings, and the detailed settings used in this work are shown in Section \ref{sec:inv_performance}.
With local samples $\{\x_i \}_{i=1}^M$, we denote the blended image as $\xblend$. 

\subsection{The DD-VAE-MCMC method}
\label{sec:implementation}
\begin{figure}[hbtp]
  \centerline{
  \begin{tabular}{c}
  \includegraphics[width=1.0\textwidth]{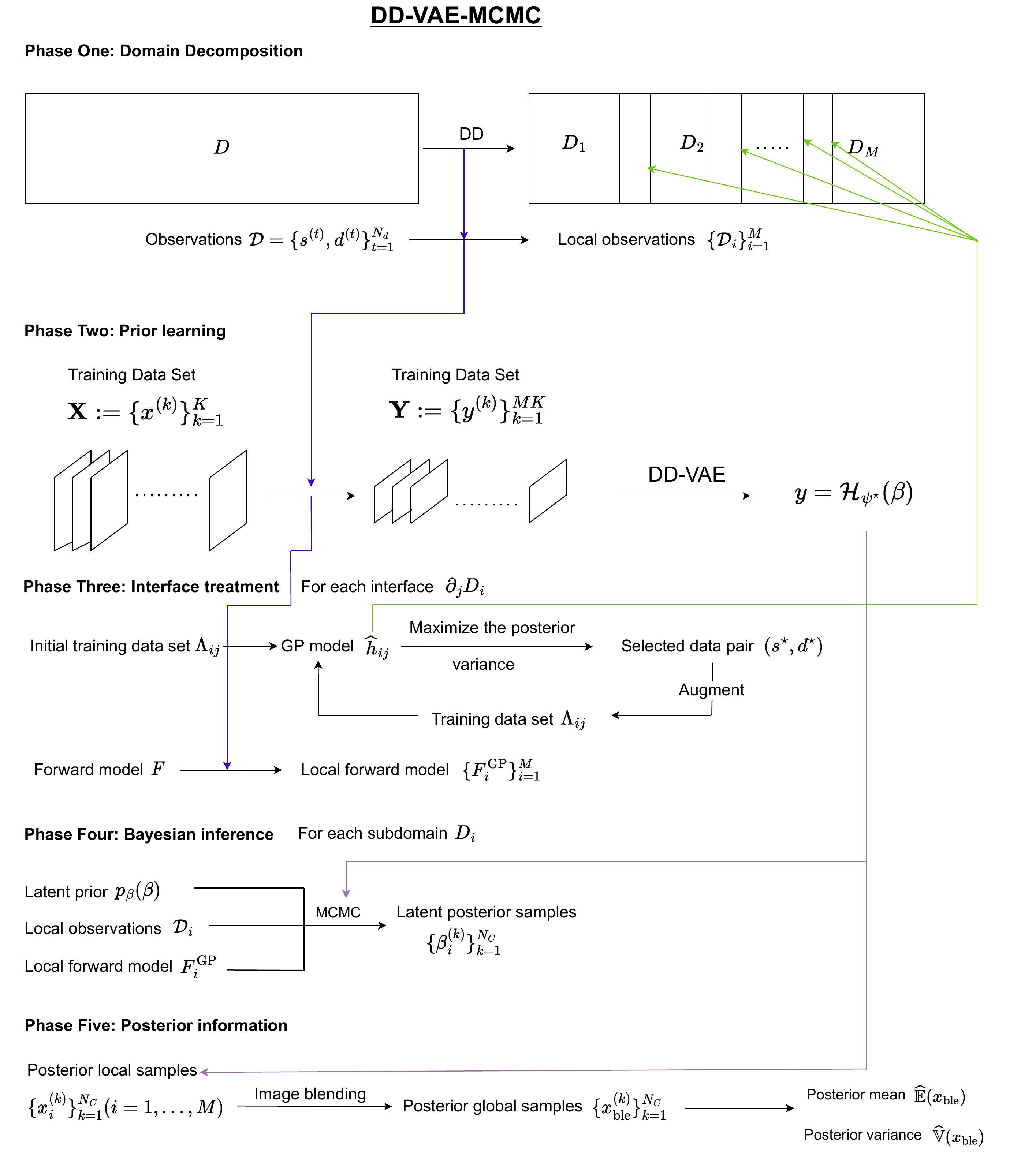}
  \end{tabular}}
  \caption{A diagram of the DD-VAE-MCMC method.}
  \label{fig:dd-vae-mcmc}
\end{figure}

In this section, we describe the detailed implementations for our DD-VAE-MCMC method.
To begin with, we divide the global domain $\D$ into $M$ overlapping but of the same size local domains $\{ \Di\}_{i=1}^M$ (see Section \ref{sec:dd_setup}).
The second phase is prior learning.
According to the strategy of the domain partition, 
we rearrange the original training data set $\DSX = \{\x^{(k)}\}_{k=1}^{\NX}$ as $\DSY = \{\y^{(k)} \}_{k=1}^{M\NX}$.
With $\DSY$, we can obtain the local generator $\lgen_{\psi^{\star}}$ using DD-VAE
(see Section \ref{sec:DD_VAE}).
The third phase is to construct interface conditions for local problems using the adaptive Gaussian process model (see Section \ref{sec:adaptive_gp}).
In the fourth phase, for each local subdomain $\D_i$, ($i=1,\ldots,M$), local latent posterior samples $\{\lla_i^{(k)}\}, k=1,\ldots,\NC$ are generated using Algorithm \ref{alg:MCMC} with local forward models $F_i^{\text{GP}}(\x_i)$ (defined through \eqref{eq:sub_problem_gp}), local observations $\bm{s}_i$ and a  prior distribution $p(\beta) = \mathcal{N}(\bm{0}, \bm{I}_{\Nlla})$. 
The last phase considers post-processing.
With local latent posterior samples $\{\lla_i^{(k)}\}_{k=1}^{\NC}$, $(i=1,\ldots,M)$, we first generate local posterior samples $\{\x_i^{(k)}\}_{k=1}^{\NC}$ for $i=1,\ldots,M$ with local generator  $\lgen_{\psi^{\star}}$.
Then the posterior global sample $\{ \xblend^{(k)}\}_{k=1}^{\NC}$ can be constructed using the Poisson image blending technique (see Section \ref{sec:blending}).
Given global posterior samples, one can compute their posterior statistical information.
A diagram for our DD-VAE-MCMC method is shown in Figure \ref{fig:dd-vae-mcmc}.
Details of our DD-VAE-MCMC method are summarized in Algorithm \ref{alg:local_vae}, where $\gptol$ is a given tolerance.

\begin{algorithm}[!ht]
  \caption{The DD-VAE-MCMC algorithm}
  \label{alg:local_vae}
  \begin{algorithmic}[1]
      \Require{Training data set $\DSX= \{ \x^{(k)} \}_{k=1}^N$, global domain $\D$, and observed data $ \calD = \{(s^{(t)}, \dobs^{(t)})\}_{t=1}^{\Nd}$.}
     \State Divide the global domain $\D$ into $M$ overlapping local domains $\{\D_i \}_{i=1}^M$ with interfaces $\partial_j \Di$ for $j\in \frakN_i$ (see the settings in Section \ref{sec:dd_setup} for details.)
      \State Divide the observation data set $\DS$ as $\{\DS_i\}_{i=1}^M$ where $\DS_i :=\{ (s^{(t)}, \dobs^{(t)})|(s^{(t)}, \dobs^{(t)})\in  \DS \text{ and } s^{(t)}\in D_i \cup \partial D_i \}$.  
      \State Rearrange the training data set $\DSX$ as $\DSY$ (see the details in Section \ref{sec:DD_VAE}).
  \State Obtain the local generator $\lgen_{\psi^{\star}}$ using Algorithm \ref{alg:vae} with altered training data set $\DSY$, the encoder $q_{\chi}(\lla|\y)$, the decoder $p_{\psi}( \y|\lla)$, the maximum epoch number $N_{e}$, and the batch size $\Nbatch$.  
  \For{each interface $\partial_j \Di$ ($(i,j)\in\frakN$)}
  \State Initialize the training set $\Lambda_{ij}$ with an arbitrary data point in $\DS_i\cup \DS_j$.
  \State Construct a finite test set $\Delta_{ij}\subset \partial_j\D_i$.
 \State Build a GP model $\widehat{h}_{ij}(\s)|\TraSetij \sim \mathcal{N}(\gpmean_{\Ndij}(\s), \sigma_{\Ndij}(\s))$ (see \eqref{eq:gp_post}).
\State Obtain the maximum posterior variance $\sigma_{\Delta_{ij}}^{\max}:= \max_{\s\in \Delta_{ij}}\sigma_{\Ndij}(\s)$.
\While{$\sigma_{\Delta_{ij}}^{\max}\ge \gptol$}
\State Find $\overline{\s} := \argmax_{\s \in \Delta_{ij}} \sigma_{\Ndij}(\s).$
\State Find $ \s^{\star}:=\argmin_{\s \in \bm{s}}\|\s - \overline{\s} \|_2\,.$
\State Update the training data set $\TraSetij= \TraSetij\cup \{(\s^{\star}), \dobs^{\star} \}$ where $\dobs^{\star}$ is the observation collected at $\s^{\star}$.
 \State Update the GP model $\widehat{h}_{ij}(\s)|\TraSetij \sim \mathcal{N}(\gpmean_{\Ndij}(\s), \sigma_{\Ndij}(\s))$ (see \eqref{eq:gp_post}).
 \State Obtain the maximum posterior variance $\sigma_{\Delta_{ij}}^{\max}:= \max_{\s\in \Delta_{ij}}\sigma_{\Ndij}(\s)$.
\EndWhile
\EndFor
\State Construct the local forward models $F_{i}^{\text{GP}}(\x_i)$ for $i=1,\ldots, M$ (see \ref{eq:sub_problem_gp}).
\For{$i=1,\ldots,M$}
\State Obtain local latent posterior samples $\{\beta_i^{(k)} \}_{k=1}^{\NC}$ using Algorithm \ref{alg:MCMC} with local model $F_{i}^{\text{GP}}(\lgen_{\psi^{\star}}(\beta))$, local observational data $\DS_i$ and the local prior $p(\lla)=\mathcal{N}(\bm{0}, \bm{I}_{\Nlla})$.
\State Obtain local posterior samples $\{\x_i^{(k)} \}_{k=1}^{\NC}$ using $\{\beta_i^{(k)} \}_{k=1}^{\NC}$ and the local generator $\lgen_{\psi^{\star}}$.
  \EndFor
  \State Using the Poisson blending technique (see Section \ref{sec:blending}) to obtain global posterior samples $\{\xblend^{(k)}\}_{k=1}^{\NC}$.
  \Ensure Global posterior samples $\{\xblend^{(k)}\}_{k=1}^{\NC}$.
  \end{algorithmic}
\end{algorithm}

\section{Numerical examples} 
\label{sec:numerical_examples}
In this section, numerical examples are conducted to  demonstrate the effectiveness of our proposed DD-VAE-MCMC method. 
Setups for the test problem are listed in Section \ref{sec:setup_num}. 
Generative properties of VAEs are discussed in Section \ref{sec:gen_num}.
Effects of the Gaussian process (GP) interface treatments are presented in Section \ref{sec:gp_num}.
Inversion results are shown in Section \ref{sec:inv_performance}.

\subsection{Setup for test problems}
\label{sec:setup_num}
Consider an elliptic PDE on a rectangular domain $D = (0,2)\times (0,1)\subset \bbR^2$ with boundary $\partial D$, 
\begin{align*}
    \nabla\cdot (-\exp(\x(\s))\nabla u(\s)) =f(\s)\,,\quad \s\in D\,,
\end{align*}
where $\s\in D$ denotes the spatial coordinate and $f$ is a known source term.
This equation describes Darcy's flow through porous media in the context of underground water. The Dirichlet boundary conditions are considered on the left and right boundaries, and the homogeneous Neumann boundary conditions are considered on the top and bottom boundaries, i.e., 
\begin{align*}
    &\exp(\x(\s))\nabla u(\s) \cdot \vec{n} = 0\,,\quad \s\in\{ (0,2)\times \{ 0\}\} \cup \{ (0,2)\times \{1\} \} \,,\\
    &u(\s) = 0\,,\quad \s\in \{ \{0\}\times (0,1)\cup \{1\}\times (0,1)  \} \,,
\end{align*}
where $\vec{n}$ denotes the normal unit vector to the Neumann boundary. The source term is set to  
\[
f(\s) = 3\exp\left(-\|\s^{sr} - \s\|_2^2\right)\,,
\]
where $\s^{sr} = [\s^{sr}_1,\s_2^{sr}]^T$ denotes the center of contaminant and it is set to $\s^{sr} = [1,0.5]^T$. 
The bilinear FEM implemented by FEniCS \cite{alnaes2015fenics} is used to discretize the problem on a $129\times 65$ grid (the number of the spatial degrees of freedom is 8385). 
The measurement noises are set to independent and identically distributed Gaussian distributions with mean zero, and the standard deviation is set to 1\% of the mean observed value.

\begin{figure}[!htp]
    \centerline{
    \begin{tabular}{c}
    \includegraphics[width=0.50\textwidth]{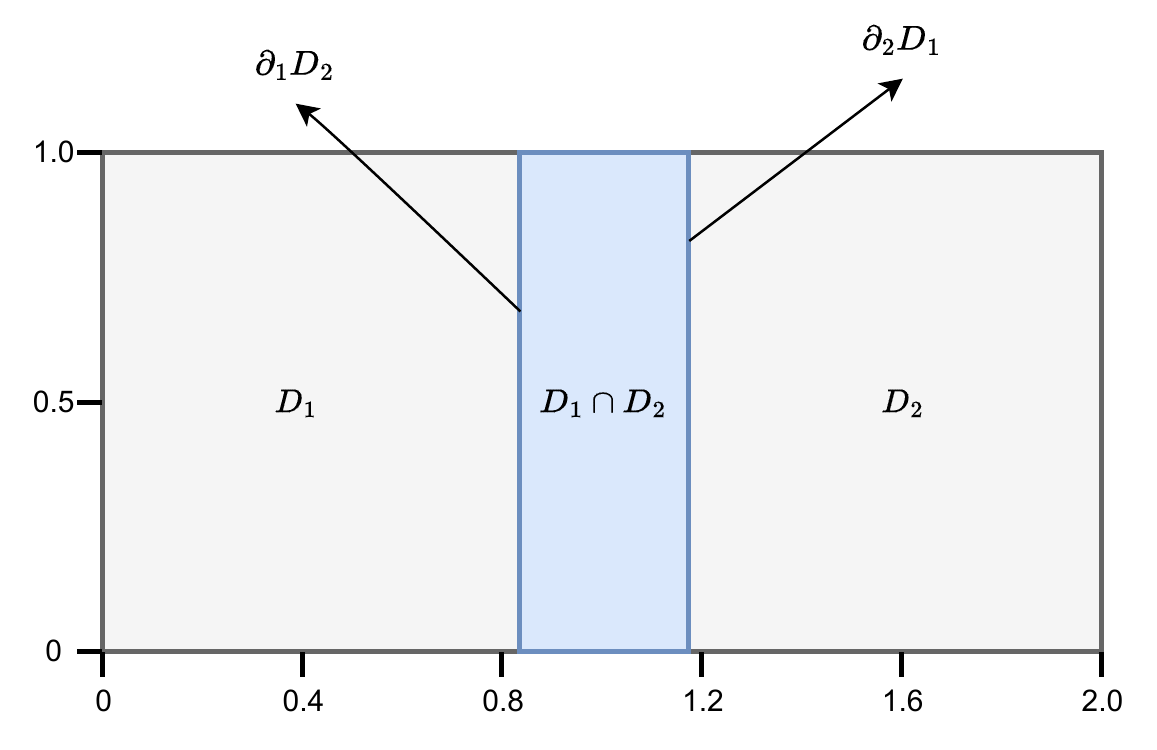}
    \end{tabular}}
    \caption{Overlapping domain decomposition with $M=2$ subdomains.}
    \label{fig:dd_demonnstrate}
\end{figure}

Let $D_1 = (0,1.1875)\times (0, 1)$, $D_2 = (0.8125, 2)\times (0, 1)$, and the interfaces are $\partial_2 \D_1 = \{1.1875\}\times (0, 1)$, $\partial_1 \D_2 = \{0.8125\}\times(0,1)$, which are shown in Figure \ref{fig:dd_demonnstrate}.
All results of this paper are obtained in Python on a workstation with a 2.20 GHz Intel(R) Xeon(R) E5-2630 GPU.
For each local subdomain, the local problem \eqref{eq:sub_problem} is discretized with the bilinear finite element method with a uniform $77\times 65$ grid (the number of the spatial degrees of freedom is 5005).

\subsection{Generative properties of VAEs}
\label{sec:gen_num}
Let $\x(\s)$ represent Gaussian random fields, i.e., $\x(\s)\sim \mathcal{GP}(\gpmean(\s), \kernel(\s,\s')$,  where $\gpmean(\cdot)$ and $\kernel(\cdot,\cdot)$ denote the mean function and the kernel function respectively.
We parameterize $\x(\s)$ via KL expansions \cite{spanos1989stochastic}. 
The covariance function $C(\s,\s'): \D\times \D\to \bbR$ is defined as 
\begin{equation*}
    C(\s,\s') = \bbE [(\x(\s,\omega) - \gpmean(\s))(\x(\s',\omega) - \gpmean(\s'))]\,. 
\end{equation*}
In this work, the covariance function is set to the exponential kernel, i.e., 
\begin{equation}
    C(s,s') = \sigma_f^2 \exp \left( -\frac{\| s-s' \|_2}{\tau}\right)\,,
    \label{eq:cov_kle}
\end{equation}
where $\tau$ is the correlation length of the covariance function and $\sigma_f$ is the standard deviation of the random field.
Suppose that $\tau$ is fixed, letting $\{ \lambda_r, \psi_r(\s)\}_{r=1}^{\infty}$ be the eigenvalues and the associated orthonormal eigenfunctions of the covariance function \eqref{eq:cov_kle},  
the Gaussian random field $\x(\s)$ can be represented via a truncated KL expansion, 
\begin{equation}
    \x(\s) \approx \gpmean(\s) + \sum_{r=1}^{\NKL} \sqrt{\lambda_r}\psi_r(\s)\kappa_r\,,
    \label{eq:kle}
\end{equation}
where $\kappa_r\sim \mathcal{N}(0,1)$, $r=1,\ldots,\NKL$, are uncorrelated standard normal random variables.
Our training data are generated by KL expansions with uncertain correlation lengths.
Letting $\sigma_f^2=0.5$ and $\gpmean(\s) = 1$, we consider two cases in this paper: a low-dimensional case where the range of the correlation length $\tau$ in \eqref{eq:kle} are taken from 1 to 2, a high-dimensional case with $\tau$ taken from 0.2 to 0.65.
To capture 95\% total variance, the number of KL terms $\NKL$ retained in the low-dimensional case is about 12$\sim$ 21, while  $\NKL$ is taken from 166 to 1462 in the high-dimensional case.
Then we discretize $\x(\s)$ with the standard finite element method on a $129\times 65$ grid. 

The number of training data set $\DSX$ for both cases is set to  $K=1\times 10^{4}$ with size $129\times 65$.
Then the augmented training data set $\DSY$ (see Section \ref{sec:DD_VAE}) contains $2\times 10^{4}$ images with size $77\times 65$.
For different domain scales and different sets of training data, it is natural to use different network structures. 
In this work, FCNNs (see Section \ref{sec:DD_VAE}) is used in the encoder and decoder models for their simplicity.  
Details settings can be found in Appendix \ref{appendix:vae_details}.
For the low-dimensional and the high-dimensional cases, G-VAE uses a Type B network structure in  Figure \ref{fig:net_struc} and DD-VAE uses a Type A network structure. 
Numerical results show that the neural network typically can be designed more shallow when considering a smaller domain.
Other hyperparameters settings can be found in Appendix \ref{appendix:vae_details}.
The dimension of the latent variable for G-VAE is set to 512, and that for DD-VAE is set to 256.
For both the low-dimensional case and the high-dimensional case, the training times for G-VAE and DD-VAE are about 56 minutes and 12 minutes, respectively. 

\begin{figure}[!htp]
    \centerline{
        \begin{tabular}{cc}
    \includegraphics[width=0.50\textwidth]{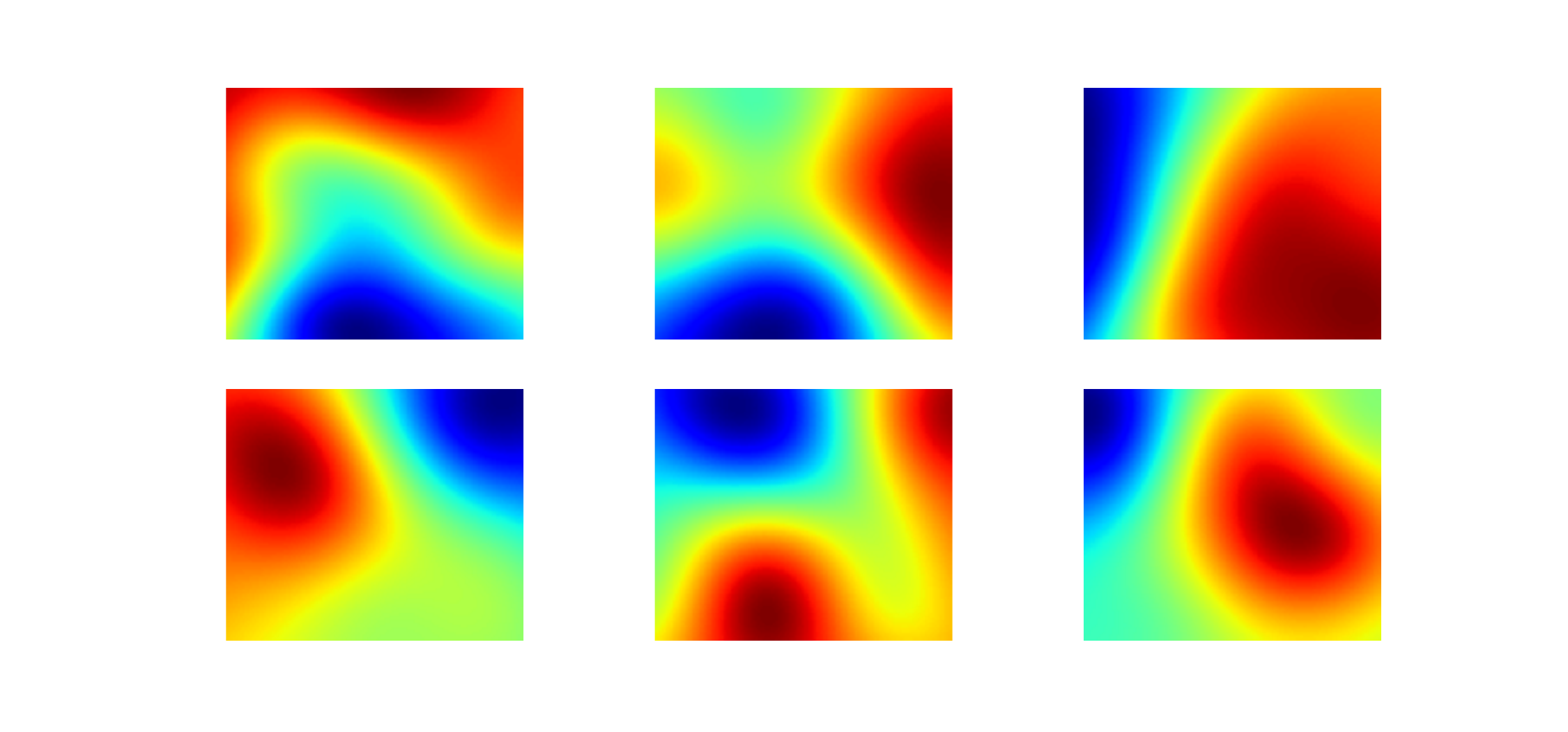}
    & 
    \includegraphics[width=0.50\textwidth]{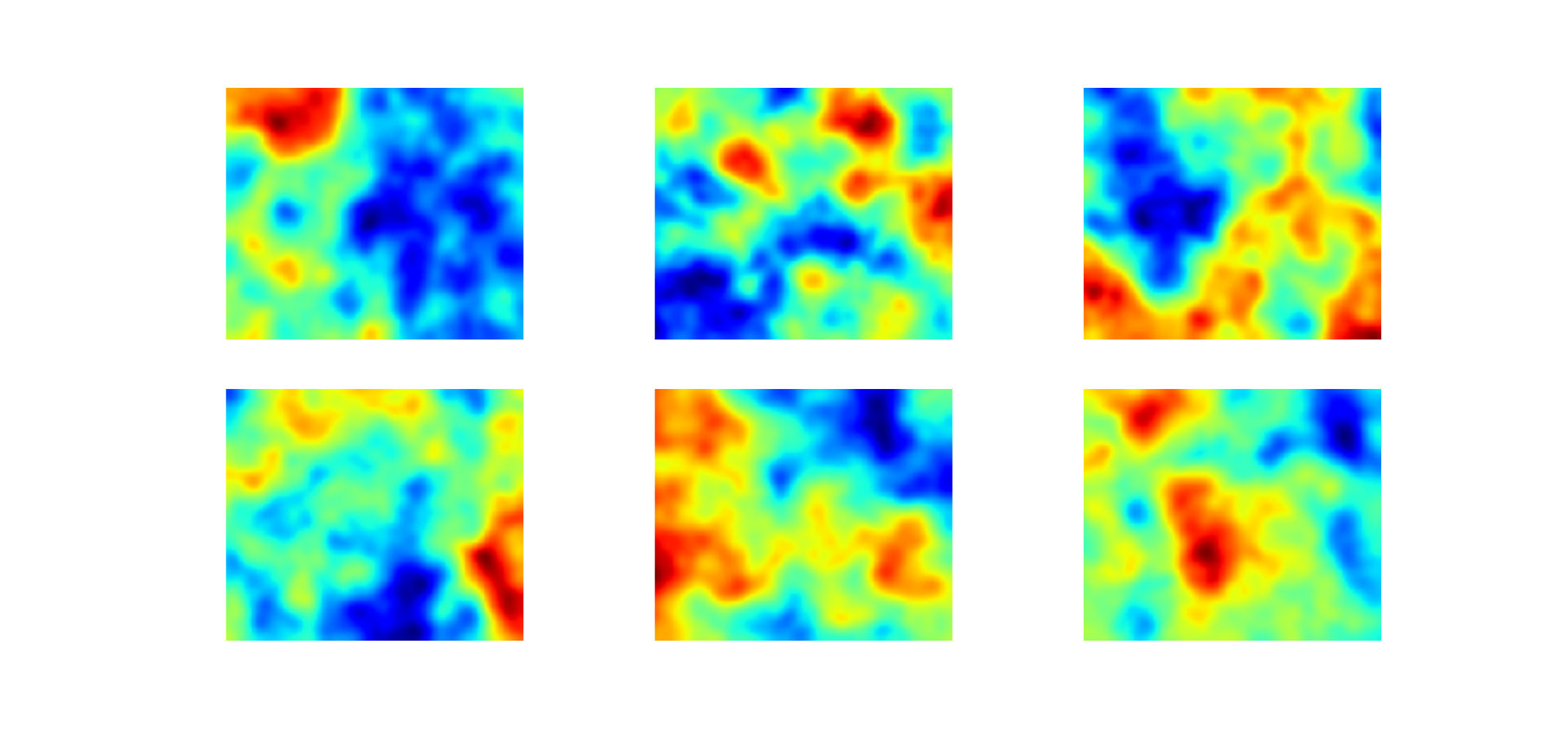}
    \\
    (a) DD-VAE (low-dimensional case)& 
    (b) DD-VAE (high-dimensional case)\\   
     \includegraphics[width=0.50\textwidth]{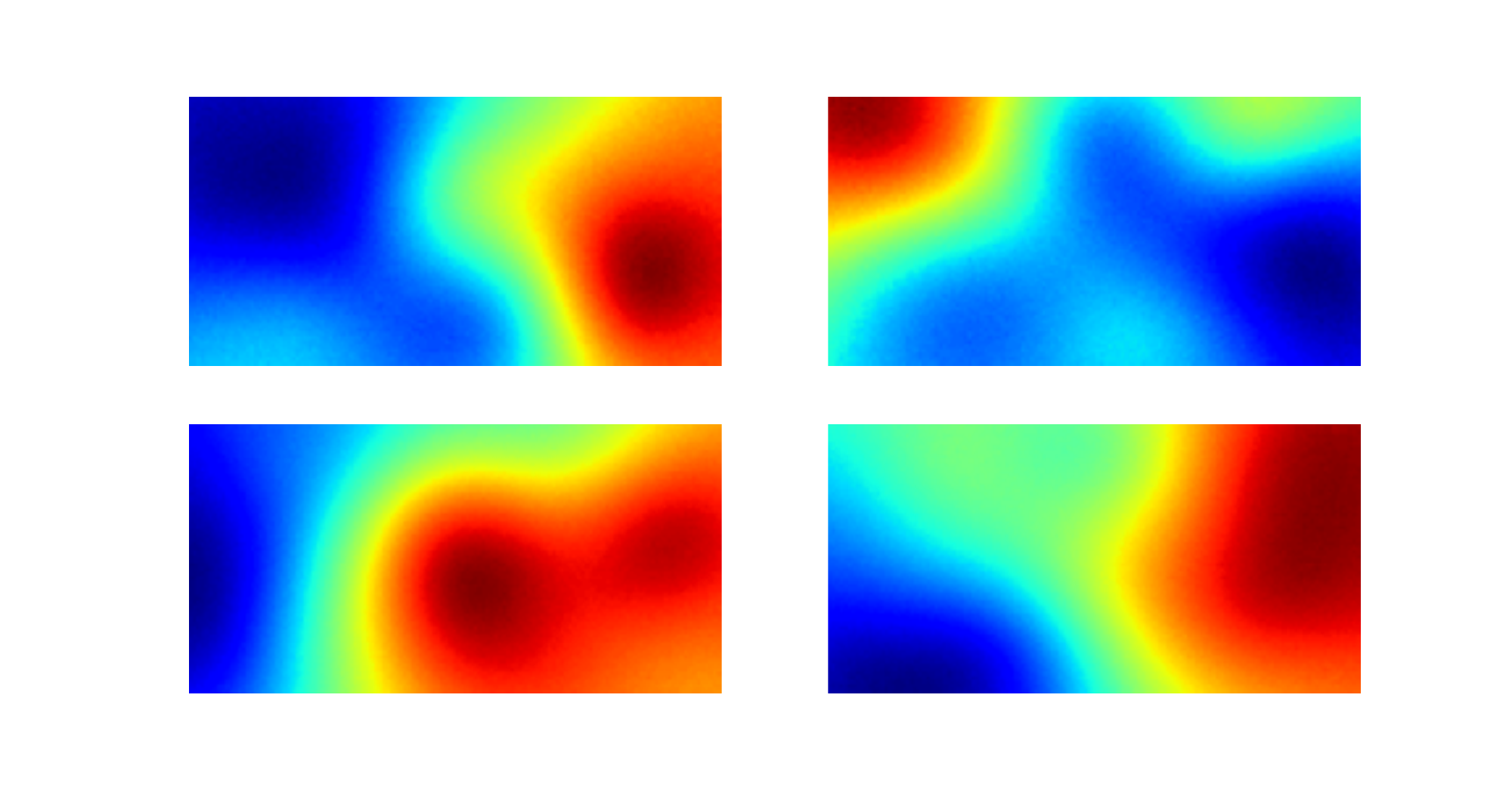} &
    \includegraphics[width=0.50\textwidth]{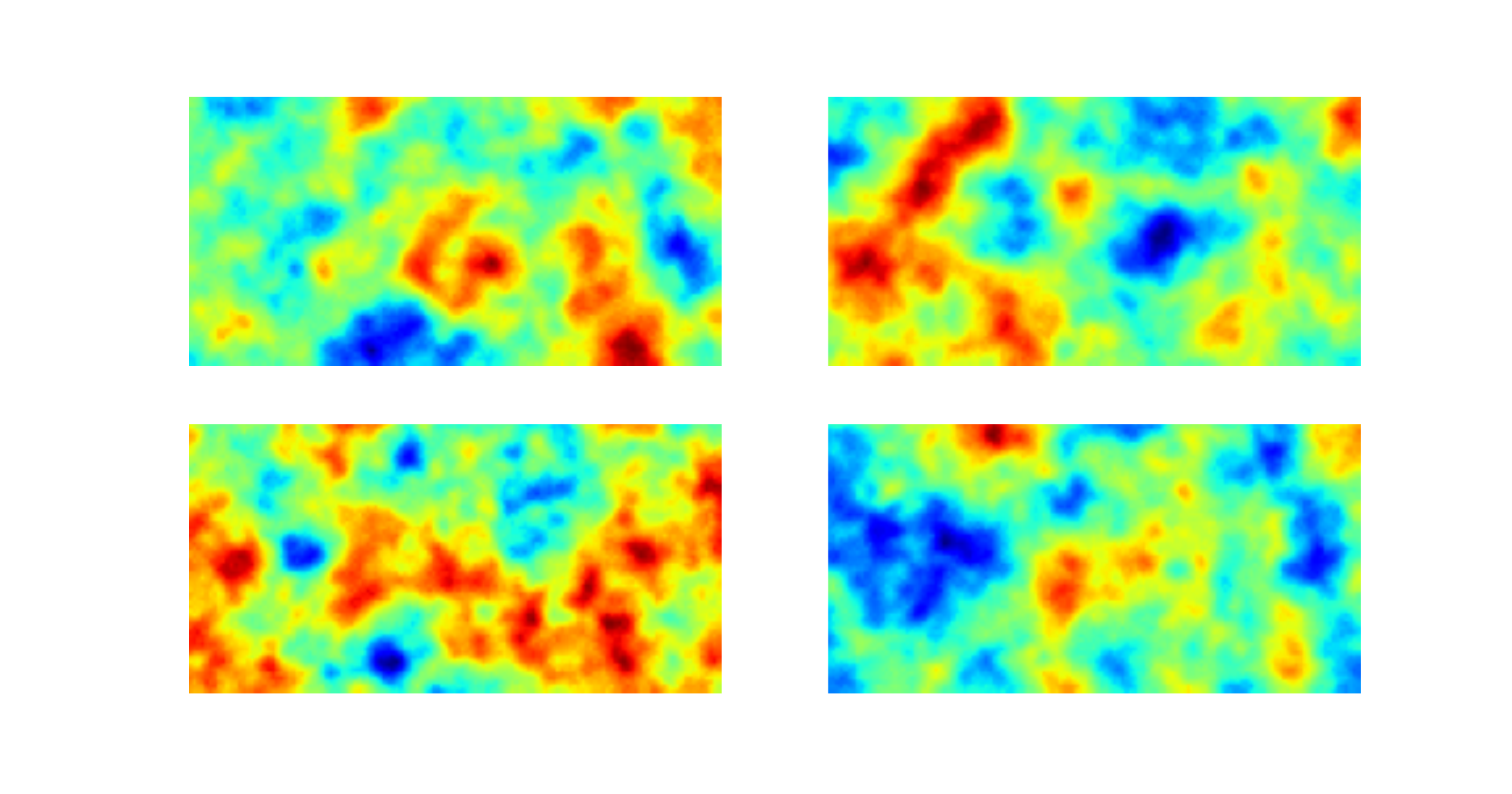}\\
    (c) G-VAE (low-dimensional case)& 
    (d) G-VAE (high-dimensional case)\\  
\end{tabular}}
\caption{Randomly generated images using DD-VAE and G-VAE.}
\label{fig:gen_results}
\end{figure}

Once the training process is completed, one can generate samples from the prior distribution by first sampling from the latent distribution and passing it through the deterministic generative model. Randomly sample $\beta$ from the standard Gaussian distribution, Figure \ref{fig:gen_results}(a) and Figure \ref{fig:gen_results}(b) show the generated images for the low-dimensional case and the high-dimensional case using D-VAE.
In contrast, Figure \ref{fig:gen_results}(c) and Figure \ref{fig:gen_results}(d) show the generated images using G-VAE. 
It can be seen that, typically, even with the higher computational cost, the generated images given by G-VAE are still more blurry compared to that given by DD-VAE.  

We introduce Fr\'echet Inception Distance (FID) \cite{heusel2017gans} to further quantify the generative property. FID measures how similar the two groups of images are based on the computer vision features of the raw images. Lower scores indicate the two groups of images are more similar or have more similar statistics   (the perfect score is 0.0, which indicates that the two groups of images are identical). 
Table \ref{tab:fid} shows FID scores for different cases and other settings for different VAE models.
We remark that since the global and local domains employ different domains, the training data used for computing the FID is inconsistent. 
It can be seen that with the same amount of computational resources, the generative property of DD-VAE significantly outperforms that of G-VAE.

\begin{table}[!htp]
	\caption{Comparisons for G-VAE and DD-VAE.}
	\centering
	\begin{tabular}{ccccccc}
		\hline
        Field Type  & Models  & \makecell[c]{Dimension of \\the input variable}  & \makecell[c]{Dimension of \\ the latent variable} & FID & \makecell[c]{Size of \\ the training data set} & \makecell[c]{Training\\time} \\
        \hline
		\multirow{2}{*}{\makecell[c]{Low-dimensional \\ Gaussian}} & G-VAE&$129\times 65$& 512& 188.006 & 10000& 56 mins \\
        & DD-VAE &$77\times 65$& 256 & 3.107& 20000& 12mins\\
        \hline
		\multirow{2}{*}{\makecell[c]{High-dimensional \\Gaussian}} & G-VAE &$129\times 65$& 512 & 393.560& 10000&56 mins\\
        & DD-VAE &$77\times 65$&256& 89.351& 20000&12mins \\
		\hline
	\end{tabular}
	\label{tab:fid}
\end{table}

\subsection{Results for the interface treatment}
\label{sec:gp_num}
For $(i,j)\in \frakN$, 
to construct a GP model for the interface function $\widehat{h}_{ij}(\s)$ over the interface $\partial_j D_i$, the test set $\Delta_{ij}$ is set to the grid points on the interfaces, where each interface has 65 grid points, and the threshold $\gptol$ for the maximum posterior variance is set to $10^{-7}$. 
We use the python package GPy \cite{gpy2014} to implement the Gaussian process regression. 
The kernel we use is the RBF kernel, i.e., $\kernel(s,\s') = \exp(-\|s-s'\|_2^2/(2\sigma_{\text{GP}}^2))$, and $\sigma_\text{GP}$ is the hyperparameter of the GP model which are automatically optimized by the training data set.

The maximum numbers of training data points required and the corresponding maximum posterior variances are shown in Table \ref{tab:gp_fitting_N_max}.
Since the target interface functions are typically smooth, Table \ref{tab:gp_fitting_N_max} shows that at most five iterations are required to reach the threshold. To assess the accuracy of the interface treatment, we compute the difference between the GP interface models and the exact interface functions associated with the truth permeability fields.
For each interface $\partial_j \D_i$, $(i,j)\in \frakN$, the relative interface error is computed through
\begin{equation}
    \reint := \| h_{ij}(\s,\x) - \widehat{h}_{ij}(\s)  \|_2/\| h_{ij}(\s,\x)\|_2
\end{equation}
where $h_{ij}(\s,\x)$ is the exact interface function defined as $h_{ij}(\s,\x):=u(\s,\x)|_{\partial_j \D_i}$ (the parameter $\x$ is associated with the truth field) and $\widehat{h}_{ij}(\s)$ is the trained GP interface model.
Moreover, for each local subdomain $\D_i$, ($i=1,2$), the relative state errors of local solutions obtained with the GP interface models are also assessed, which are computed through
\[
    \epsilon_{i}^{\text{(state)}}:=\|u_{i}^{\text{GP}}(\s,\x_i) -u_i(\s,\x_i) \|_2/\|u_i(\s,\x_i) \|_2\,,
\]
where $u_i^{\text{GP}}(\s,\x_i)$ is the local solution defined in \eqref{eq:sub_problem_gp}, $u_i(\s,\x_i)$ is the exact local solution which is defined in \eqref{eq:sub_problem}. Table \ref{tab:gp_error} shows the relative interface errors and state errors, and it can be seen that these errors are all small.

\begin{table}[!htp]
	\caption{Maximum number of training data points and the corresponding maximum posterior variances for the two test cases. }
	\centering
	\begin{tabular}{c|c c c}
		\hline
		 Field cases & \makecell[c]{Gaussian field \\low-dimensional case}  &\makecell[c]{Gaussian field \\high-dimensional case}  &  \\
		\hline
         $|\Lambda_{12}|$ &2 &3 &    \\  
        $|\Lambda_{21}|$ & 5&3   & \\
		\hline
		$\sigma_{\Delta_{12}}^{\max} $&$9.0460\times 10^{-8}$& $3.4867\times 10^{-10}$   & \\  
        $\sigma_{\Delta_{21}}^{\max} $ &$2.1025\times 10^{-10}$& $ 1.856\times 10^{-10}$   & \\
		\hline
	\end{tabular}
	\label{tab:gp_fitting_N_max}
\end{table}

\begin{table}[!htp]
	\caption{Relative errors for different interfaces.}
	\centering
	\begin{tabular}{c|c|ccc}
		\hline
        &Field cases & \makecell[c]{Gaussian field \\low-dimensional case}  &\makecell[c]{Gaussian field \\high-dimensional case}  &  \\
		\hline
		\multirow{2}*{$\reint$} &
        $\partial_2 \D_1$ & $1.6721\times 10^{-2}$ & $5.2275 \times 10^{-2}$ & \\ 
		~ &  $\partial_1 \D_2$ & $3.3415\times 10^{-2}$ & $1.1325 \times 10^{-2}$  & \\  
\hline
\multirow{2}*{$\epsilon_{i}^{\text{(state)}}$} & $\D_1$ &$1.8193\times 10^{-2}$& $ 2.8392\times 10^{-2}$   & \\
~ &$\D_2$& $8.3981\times 10^{-3}$& $7.1597 \times 10^{-3}$ &  \\
\hline
	\end{tabular}
	\label{tab:gp_error}
\end{table}

\subsection{Inversion performance: GRFs with uncertain correlation lengths}
\label{sec:inv_performance}
 
\begin{figure}[!htp]
    \centerline{
        \begin{tabular}{cc}
    \includegraphics[width=0.50\textwidth]{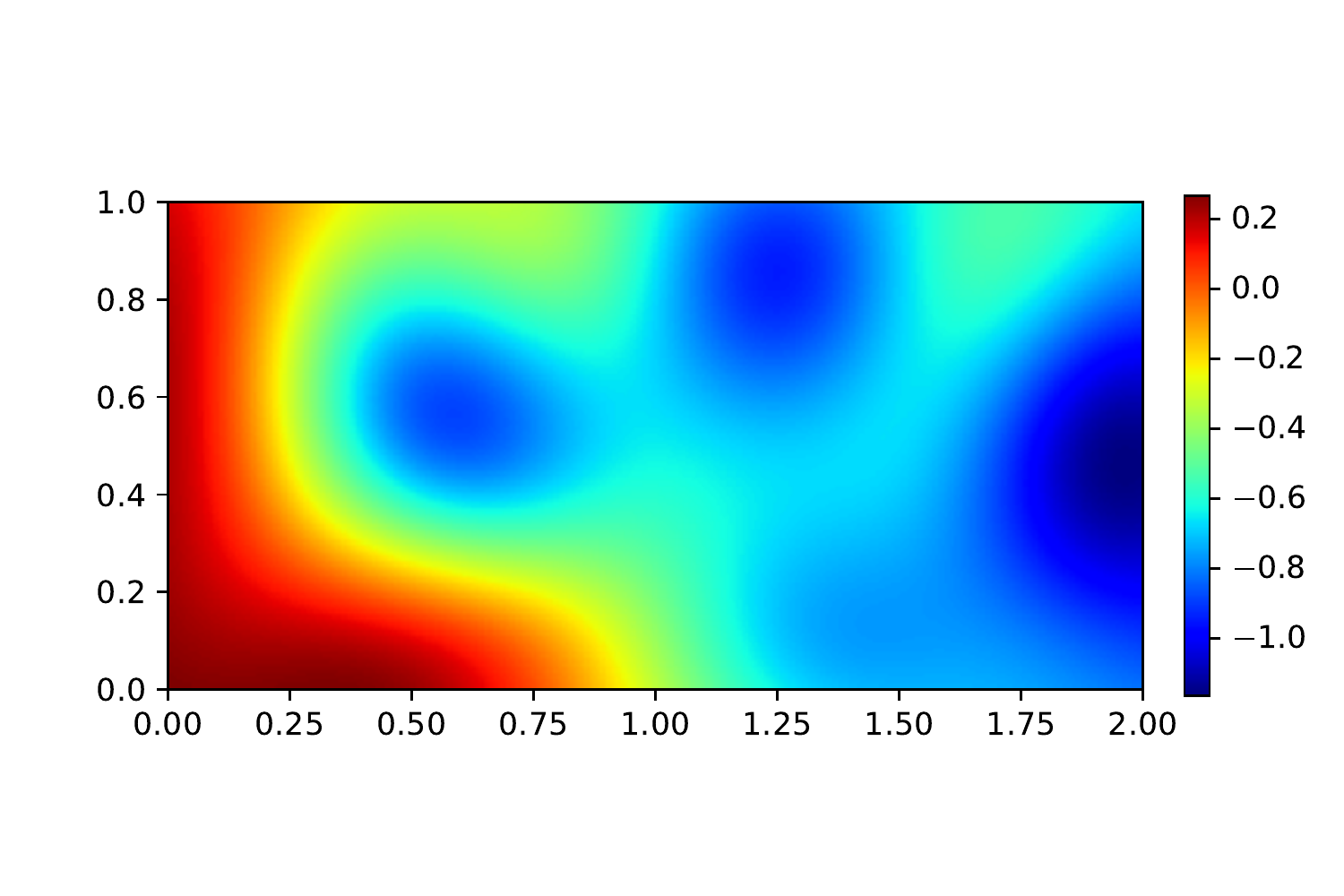}
    & 
    \includegraphics[width=0.50\textwidth]{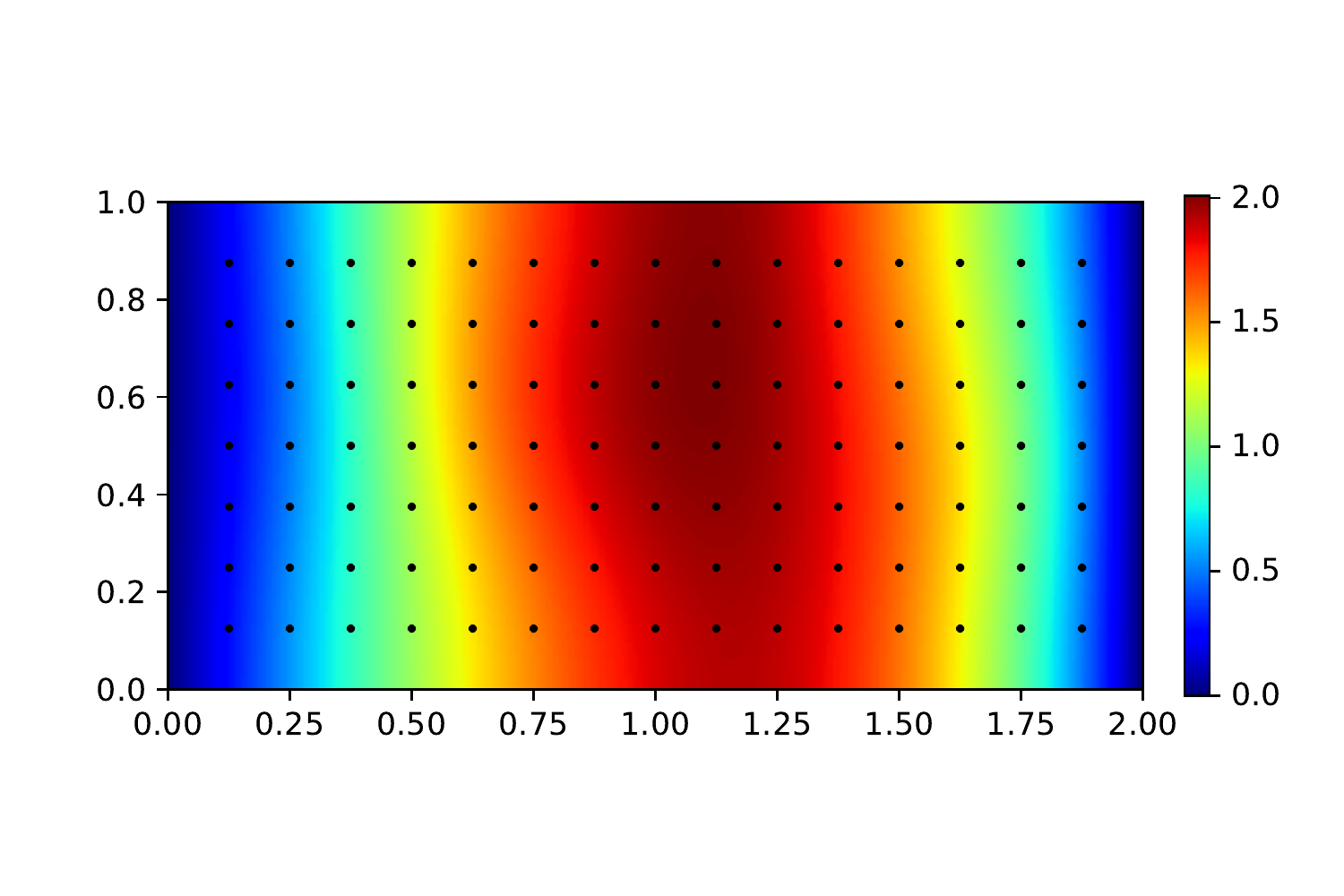}\\
    (a) The truth log permeability field. & 
    (b) The truth pressure field and sensors. \\
\end{tabular}}
\caption{Test problem setup (low-dimensional case).}
\label{fig:setup_low}
\end{figure}

\begin{figure}[!htp]
    \centerline{
        \begin{tabular}{cc}
    \includegraphics[width=0.50\textwidth]{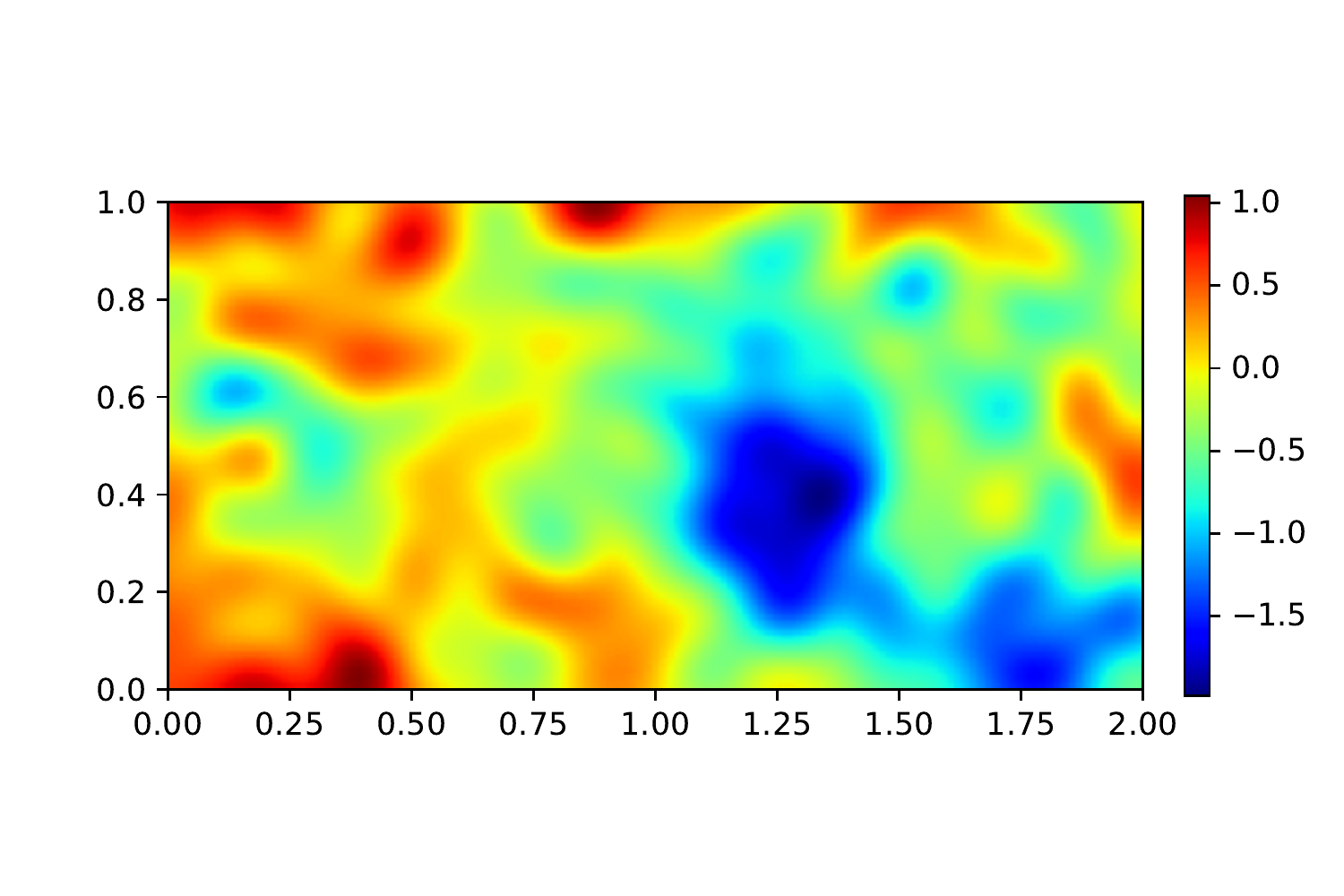}
    & 
    \includegraphics[width=0.50\textwidth]{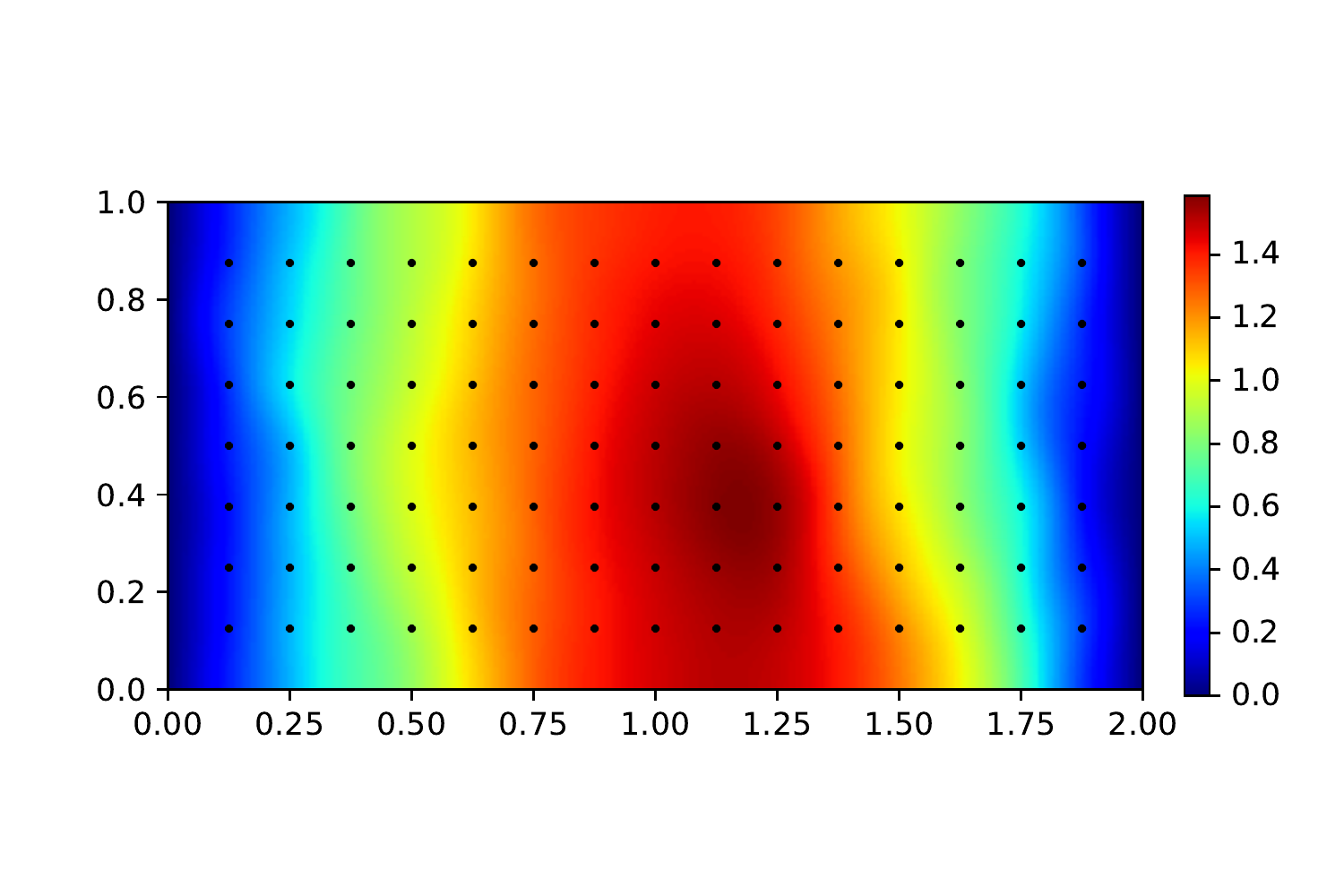}\\
    (a) The truth log permeability field. & 
    (b) The truth pressure field and sensors. \\
\end{tabular}}
\caption{Test problem setup (high-dimensional case).}
\label{fig:setup_high}
\end{figure}

The truth fields for the low-dimensional case and the high-dimensional case are shown in Figure \ref{fig:setup_low}(a) and Figure \ref{fig:setup_high}(a).
Both of them are not included in the training data. 
The corresponding pressure fields are shown in Figure \ref{fig:setup_low}(b) and Figure \ref{fig:setup_high}(b), and black dots denote the locations of sensors. 
Solving the linear system associated with the global model for the low-dimensional case and the high-dimensional case takes about $1.3529$ seconds. 
Moreover, the local models with two cases can be solved in $0.3299$ seconds.  
The ratios of the global and local model evaluations are about 4.1009. 
Therefore, defining the computational cost to conduct a local forward model evaluation as one cost unit, the cost for each global model evaluation is roughly $4$ cost units.  
The acceptance rate of MCMC is defined as the ratio of accepted samples divided by the total sample size.
The step sizes $\gamma$ (see \eqref{eq:pCN_proposal}) for both DD-VAE-MCMC and G-VAE-MCMC are carefully tuned such that the acceptance rates are appropriate.
The length of the Markov chain in G-VAE-MCMC is four times of that in DD-VAE-MCMC to ensure the computational cost is approximately equal.
Step sizes, the length of the Markov chain, and the acceptance rates for the low and high-dimensional cases are summarized in Table \ref{tab:mcmc_low} and Table \ref{tab:mcmc_high}, respectively.

The settings we use for the Poisson image blending are, for two overlapping images $\x_1\in \D_1$ and $\x_2\in \D_2$, $\Omega$ is set to $\Omega:=\D_1\cap \D_2$, we define the known scalar function as 
\[
  I^{\star}(s) = \begin{cases}
  \x_1(\s)\,, \quad  \s\in D_1/\Omega\,,\\
  \x_2(\s)\,,\quad \s \in D_2/\Omega\,.
  \end{cases}
\]
The source image $B$ is set to 
\[
    B(s) = \frac{1}{2} (\x_1(\s) + \x_2(\s))\,,\quad \s \in \Omega\,,
\]
and the obtained blended image $I$ is denoted as $\xblend$.
For comparison, the \textit{stitched field} $\xstitch$ of images $\x_1$ and $\x_2$ is defined as follows
\begin{equation}
    \xstitch = \sum_{i=1}^M \x_i \cdot \mathbb{I}_{D_i}(\s)\,,   
    \label{eq:stitch}
\end{equation}
where the indicator function $\mathbb{I}_{D_i}(\s)$ is defined as 
\[
  \mathbb{I}_{D_i}(\s):=\begin{cases}
    0\,,\quad \text{if $\s\notin \D_i$ }\,,\\
     1\,,\quad \text{if $\s\in \D_i/\left( \cup_{j\in\frakN_i} (\D_i\cup \D_j) \right)$ }\,,\\
     \frac{1}{|\frakN_i|}\,,\quad \text{if $\s\in\cup_{j\in\frakN_i} (\D_i\cup \D_j)$ }\,.
  \end{cases}
\]
In this work, the stitched field is 
\[
\xstitch = \begin{cases}
    x_1(\s)\,,\quad \s \in D_1/\Omega\,,\\
    \frac{1}{2}\left(x_1(\s) + x_2(\s)\right)\,,\quad \s \in \Omega\,,\\
    x_2(\s)\,,\quad \s \in D_2/\Omega\,.
\end{cases}
\]

To quantify the uncertainty of the generated posterior samples, we compute the mean and the variance estimates of the permeability fields as follows. 
Given posterior samples $\{ \x^{(k)}\}_{k=1}^K$ defined in the global domain $\D$, the posterior mean and variance estimates are computed through 
\begin{align}
    &\widehat{\bbE}(\x) := \frac{1}{K}\sum_{k=1}^K \x^{(k)}\,,
    \label{eq:post_mean}\\
    &\widehat{\bbV}(\x) := \frac{1}{K}\sum_{k=1}^K \left[ \x^{(k)} - \widehat{\bbE}(\x)\right]^2\,.
    \label{eq:post_var}
\end{align}
We denote the samples generated by G-VAE-MCMC as $\{\xg^{(k)}\}_{k=1}^{\NC}$.
The posterior mean $\widehat{\bbE}(\xg)$ and variance $\widehat{\bbV}(\xg)$ of G-VAE-MCMC can be computed by putting samples $\{\xg^{(k)}\}_{k=1}^{\NC}$ into \eqref{eq:post_mean} and \eqref{eq:post_var}.
The posterior mean $\widehat{\bbE}(\xblend)$ and variance $\widehat{\bbV}(\xblend)$ for DD-VAE-MCMC can be computed by putting samples $\{\xblend^{(k)}\}_{k=1}^{\NC}$ generated by Algorithm \ref{alg:local_vae} into \eqref{eq:post_mean} and \eqref{eq:post_var}.
The posterior stitched samples $\{\xstitch^{(k)}\}_{k=1}^{\NC}$ are computed by putting local posterior samples $\{\x_i^{(k)}\}_{k=1}^{\NC}$ ($i=1,2$) in \eqref{eq:stitch}.
Then the posterior mean $\widehat{\bbE}(\xstitch)$ and variance $\widehat{\bbV}(\xstitch)$ for the stitched field can be computed by putting samples $\{\xstitch^{(k)}\}_{k=1}^{\NC}$ into \eqref{eq:post_mean} and \eqref{eq:post_var}. 

To access the accuracy of the estimated posterior mean fields, relative errors for three cases are defined as 
\begin{align}
    &\epsilon_{\text{sti}}:= \|\widehat{\bbE}(\xstitch) -\xtruth\|_2/ \| \xtruth\|_2\,, \\
    &\epsilon_{\text{ble}}:=  \|\widehat{\bbE}(\xblend)-\xtruth \|_2/ \| \xtruth\|_2\,,\\
    &\epsilon_{\text{g}}:= \| \widehat{\bbE}(\xg)-\xtruth \|_2/ \| \xtruth\|_2 \,,
\end{align}
where $\xtruth$ is the truth.
Table \ref{tab:compare_err} shows the relative errors in the mean estimates for the two test problems. 
Our DD-VAE-MCMC can typically give smaller errors. Stitched fields give larger relative errors, and the corresponding images have visible seams. 

\begin{figure}[!htp]
    \centerline{
        \begin{tabular}{cc}
    \includegraphics[width=0.50\textwidth]{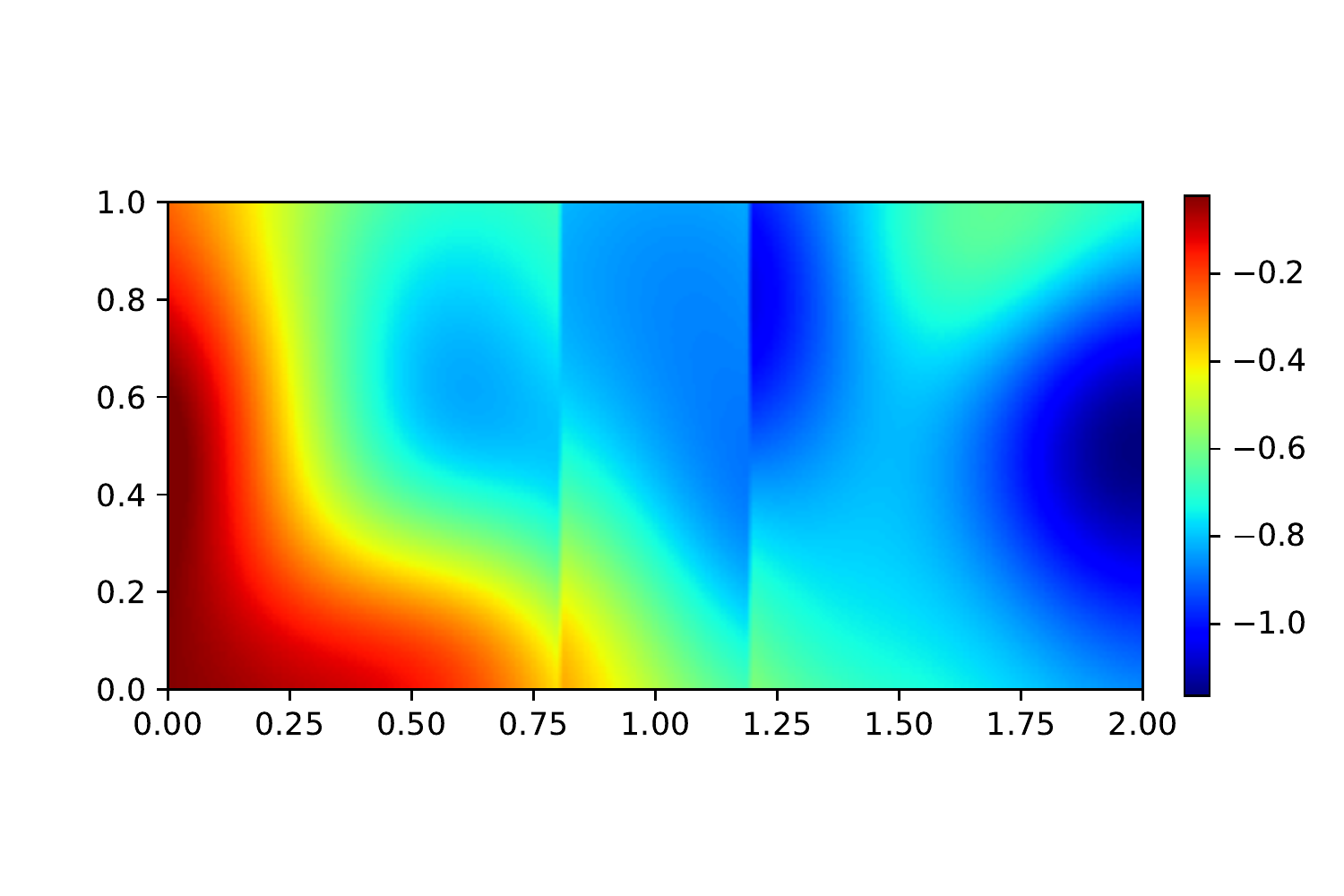}
    & 
    \includegraphics[width=0.50\textwidth]{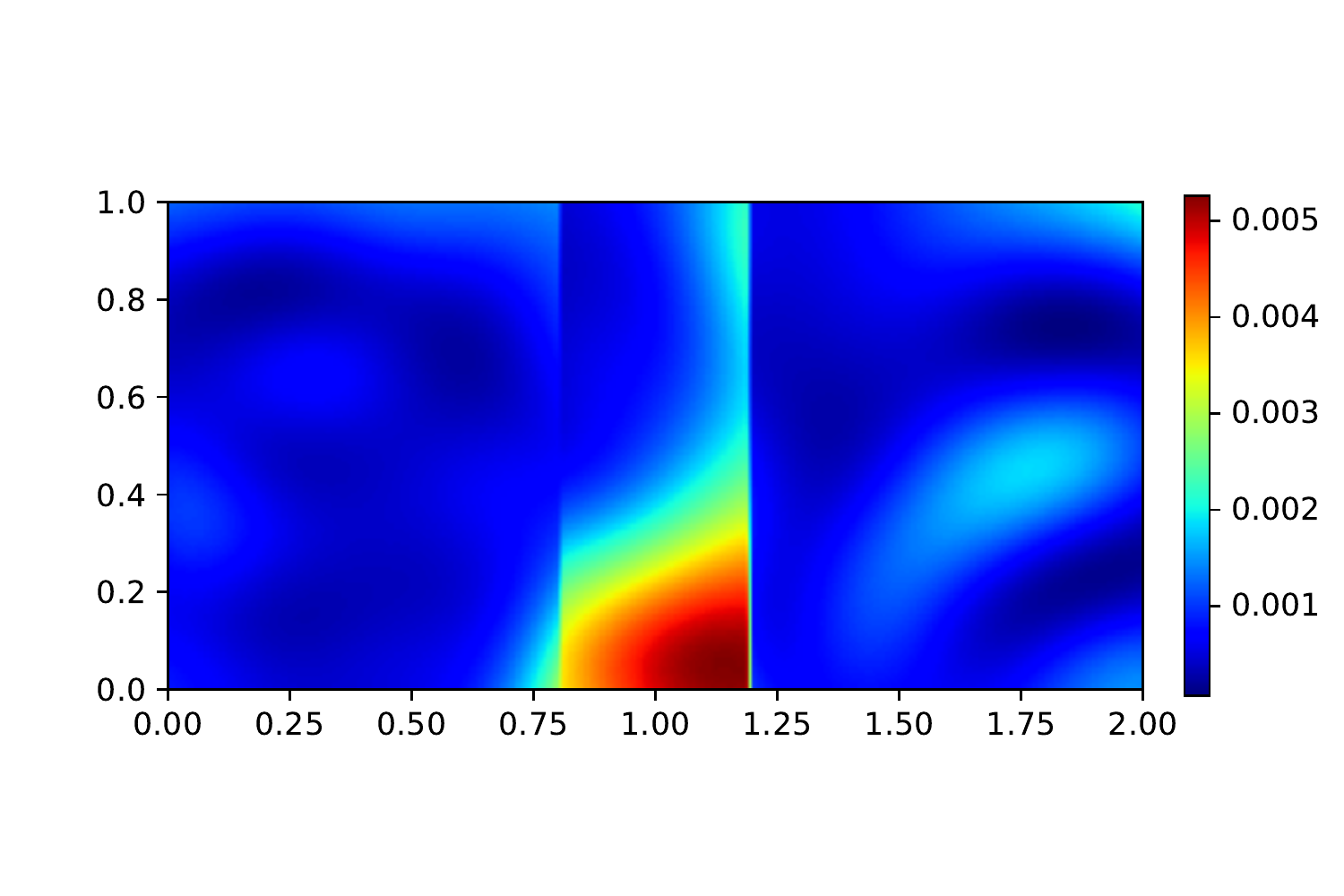}\\
    (a) Mean $\widehat{\bbE}(\xstitch)$, the stitched filed. & 
    (b) Variance $\widehat{\bbV}(\xstitch)$, the stitched filed. \\
    \includegraphics[width=0.50\textwidth]{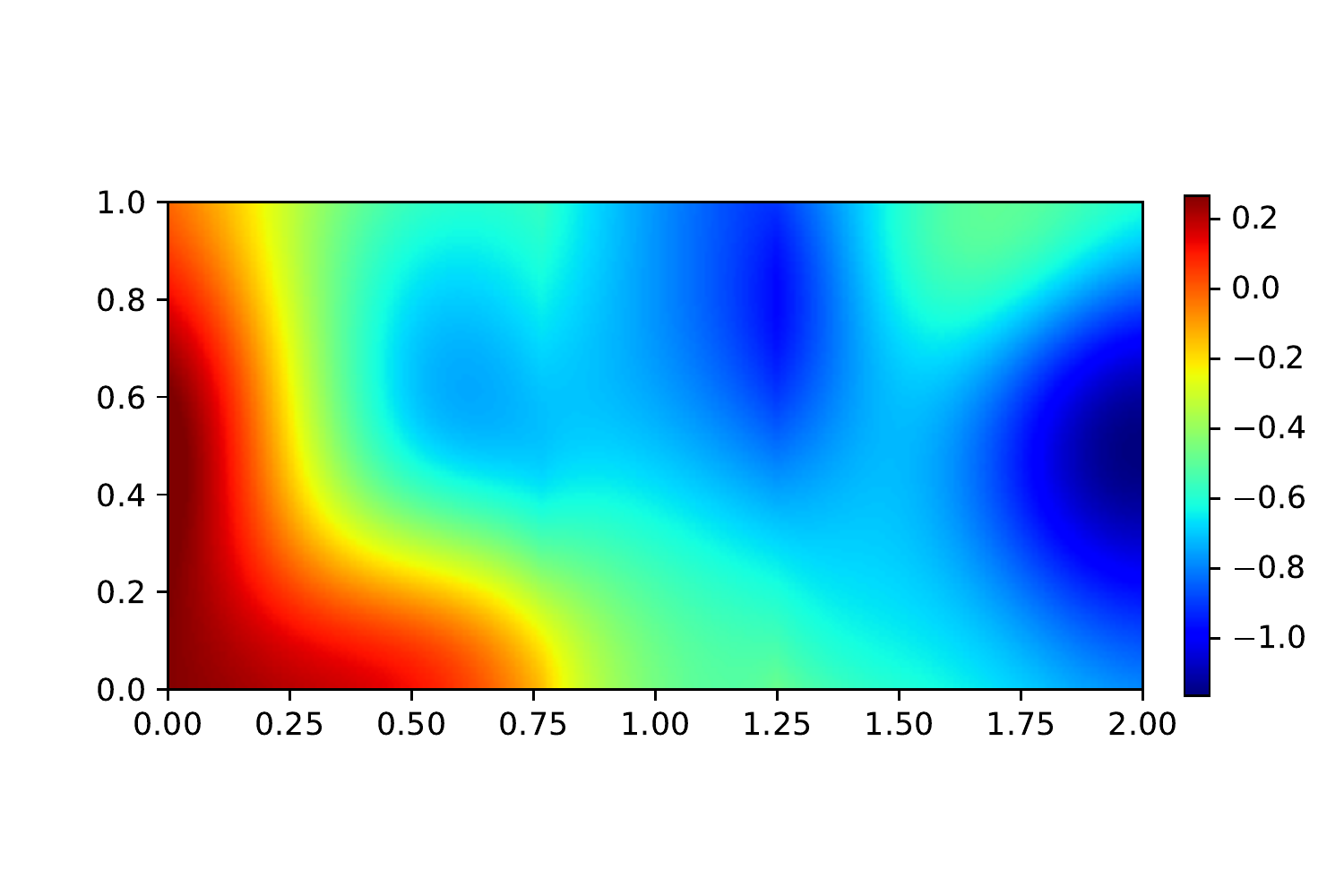}
    & 
    \includegraphics[width=0.50\textwidth]{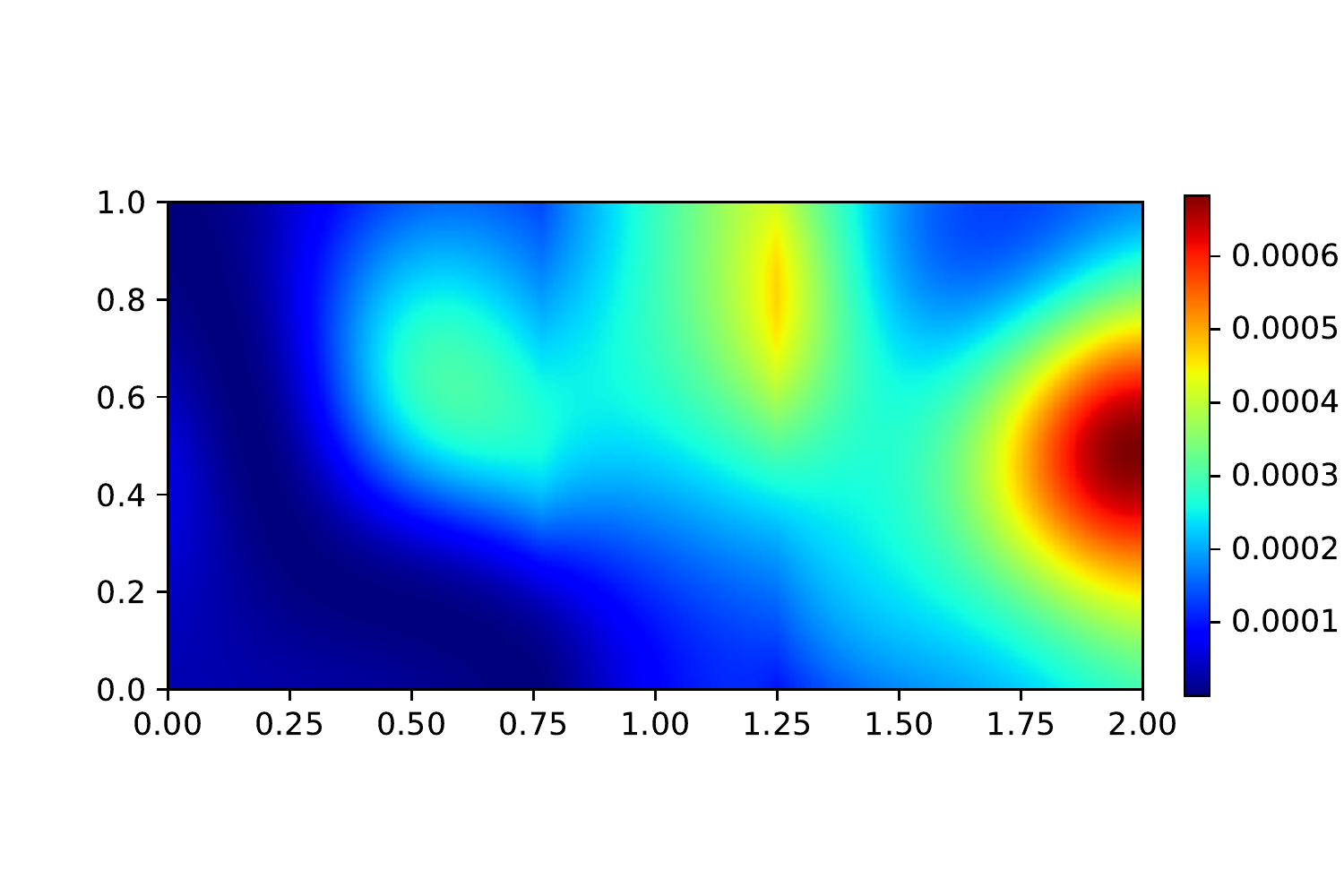}\\
    (c) Mean $\widehat{\bbE}(\xblend)$, DD-VAE-MCMC. & 
    (d) Variance $\widehat{\bbV}(\xblend)$, DD-VAE-MCMC. \\
    \includegraphics[width=0.50\textwidth]{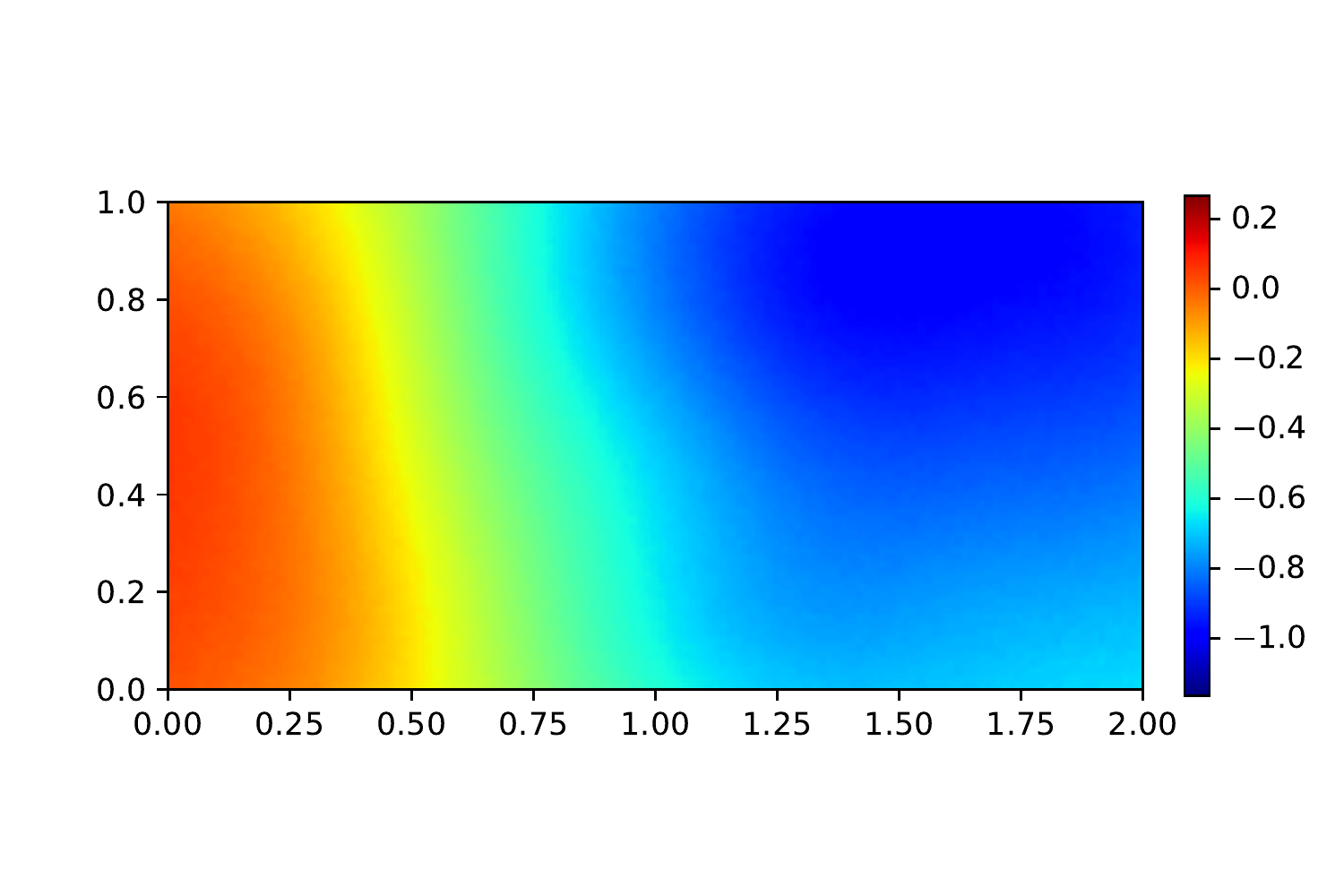}
    & 
    \includegraphics[width=0.50\textwidth]{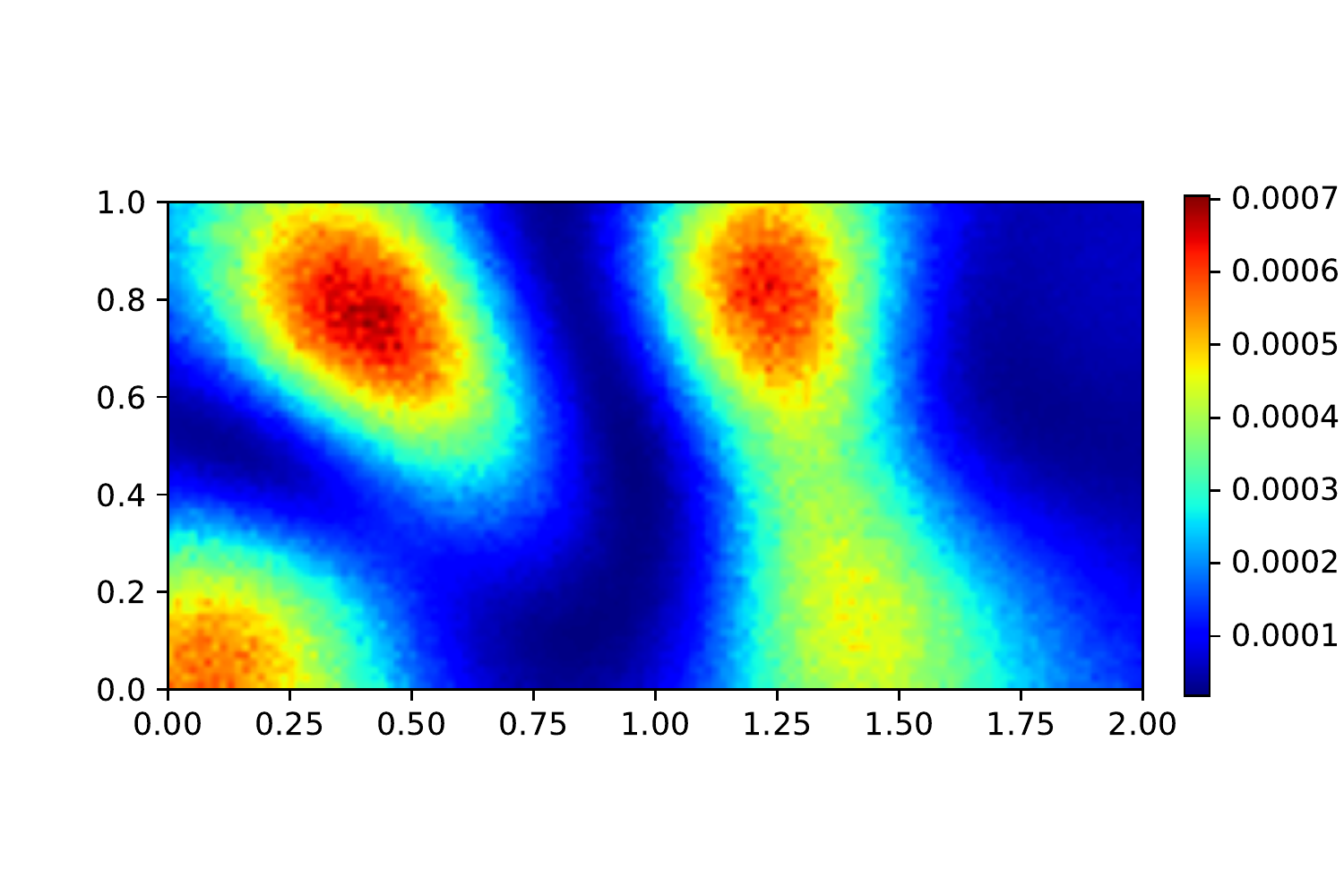}\\
    (e) Mean $\widehat{\bbE}(\xg)$, G-VAE-MCMC. & 
    (f) Variance $\widehat{\bbV}(\xg)$, G-VAE-MCMC. \\
\end{tabular}}
\caption{Estimated mean and variance fields (low-dimensional case).}
\label{fig:compare_low}
\end{figure}

\begin{figure}[!htp]
    \centerline{
        \begin{tabular}{cc}
    \includegraphics[width=0.50\textwidth]{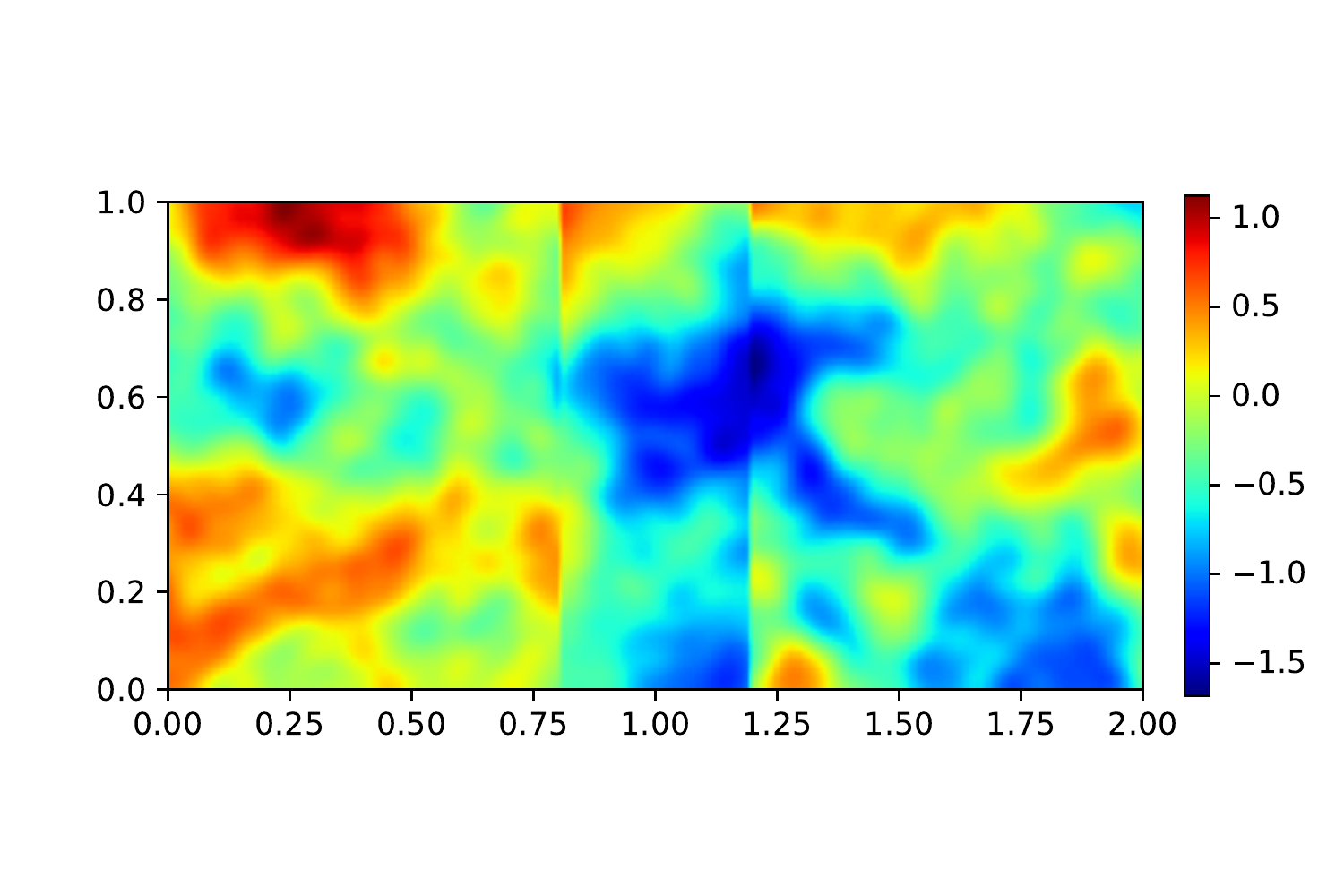}
    & 
    \includegraphics[width=0.50\textwidth]{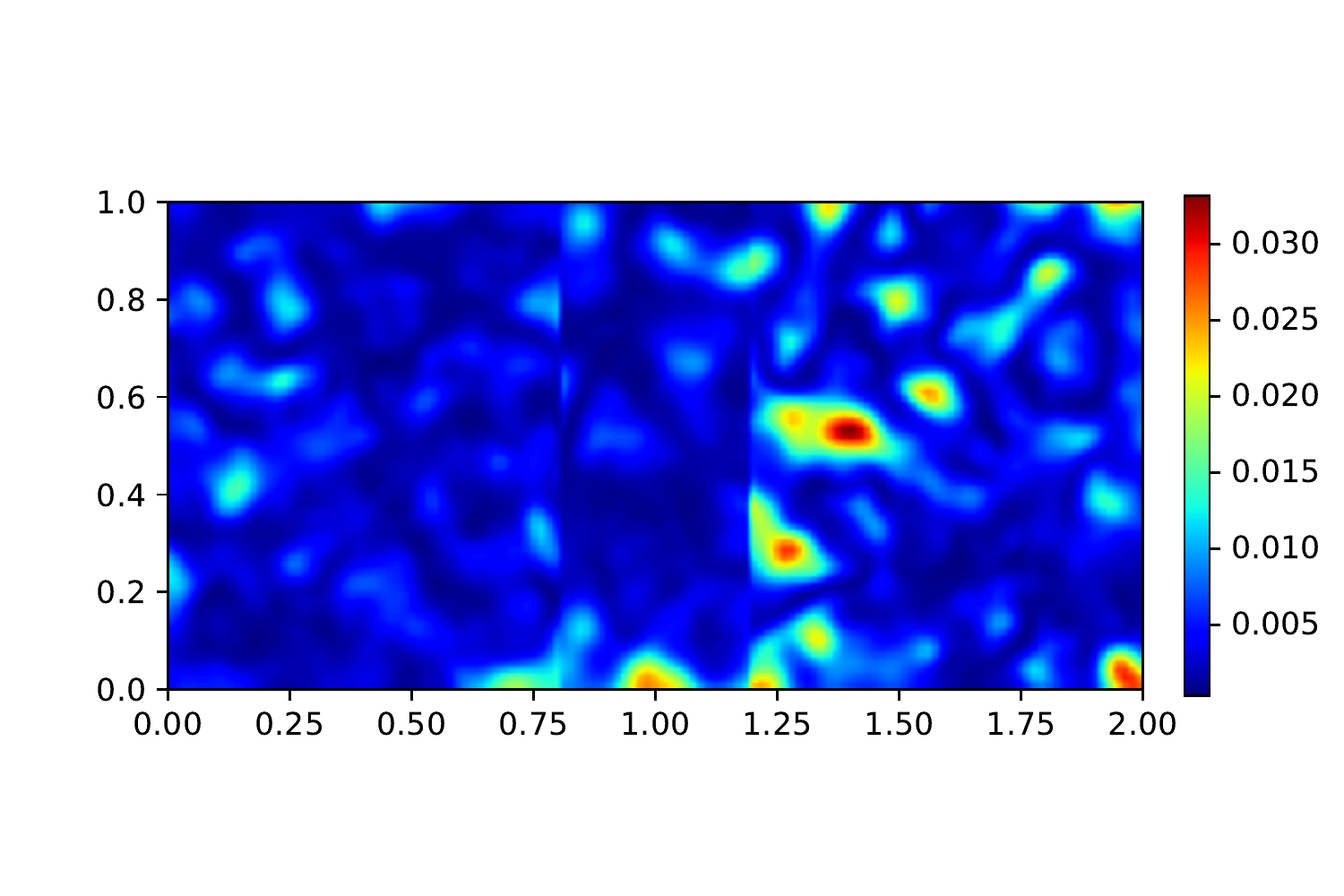}\\
    (a) Mean $\widehat{\bbE}(\xstitch)$, the stitched filed. & 
    (b) Variance $\widehat{\bbV}(\xstitch)$, the stitched filed. \\
    \includegraphics[width=0.50\textwidth]{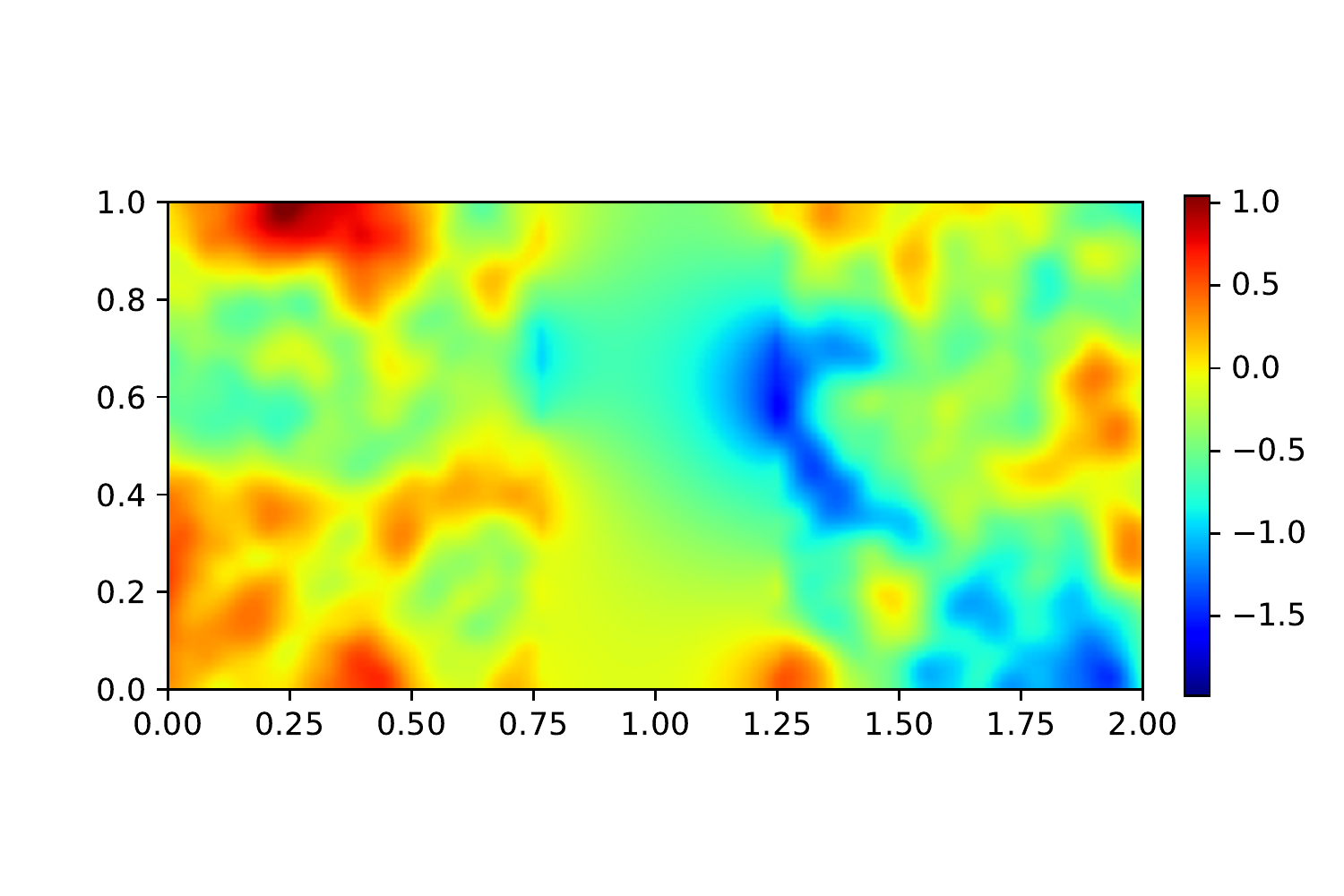}
    & 
    \includegraphics[width=0.50\textwidth]{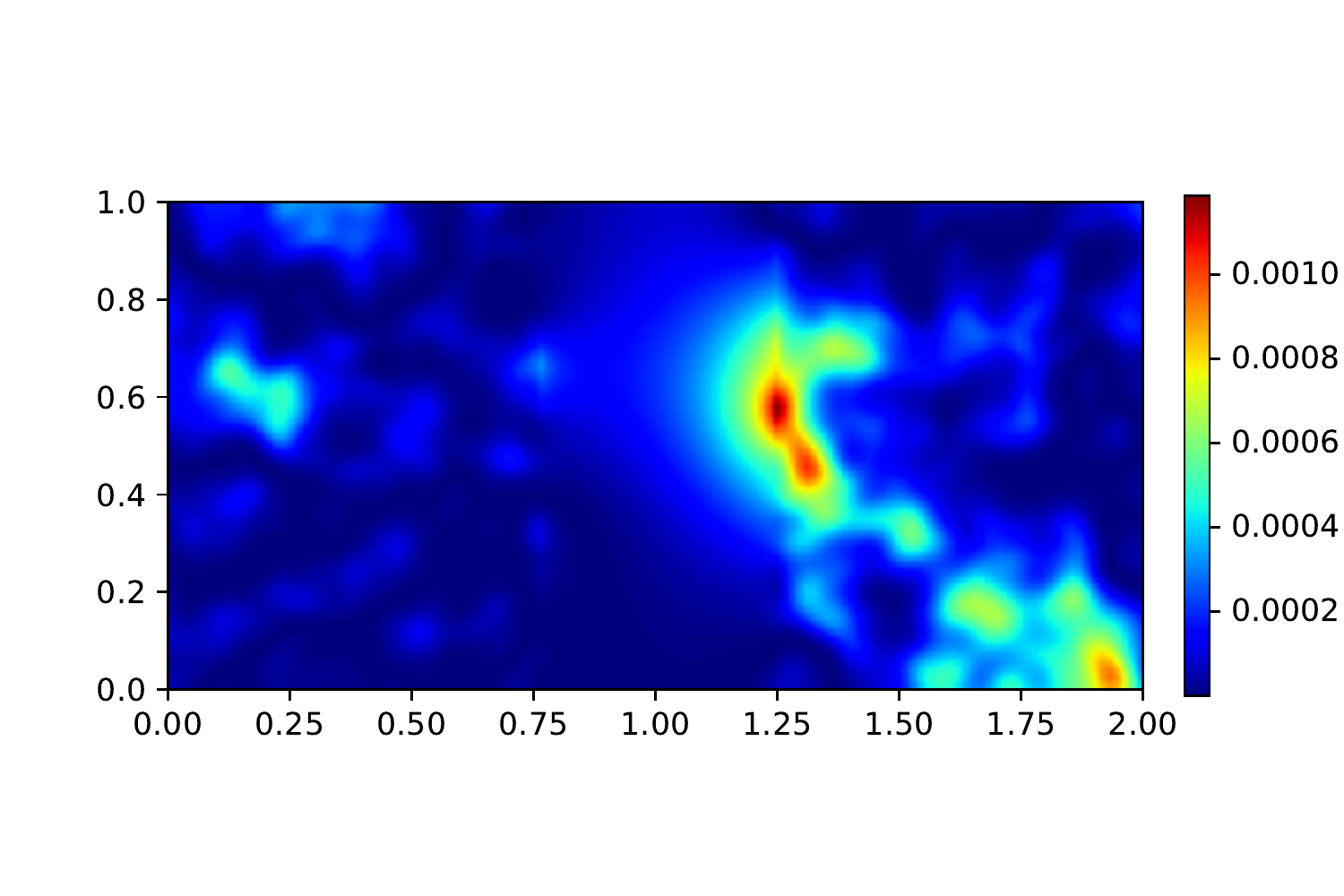}\\
    (c) Mean $\widehat{\bbE}(\xblend)$, DD-VAE-MCMC. & 
    (d) Variance $\widehat{\bbV}(\xblend)$, DD-VAE-MCMC. \\
    \includegraphics[width=0.50\textwidth]{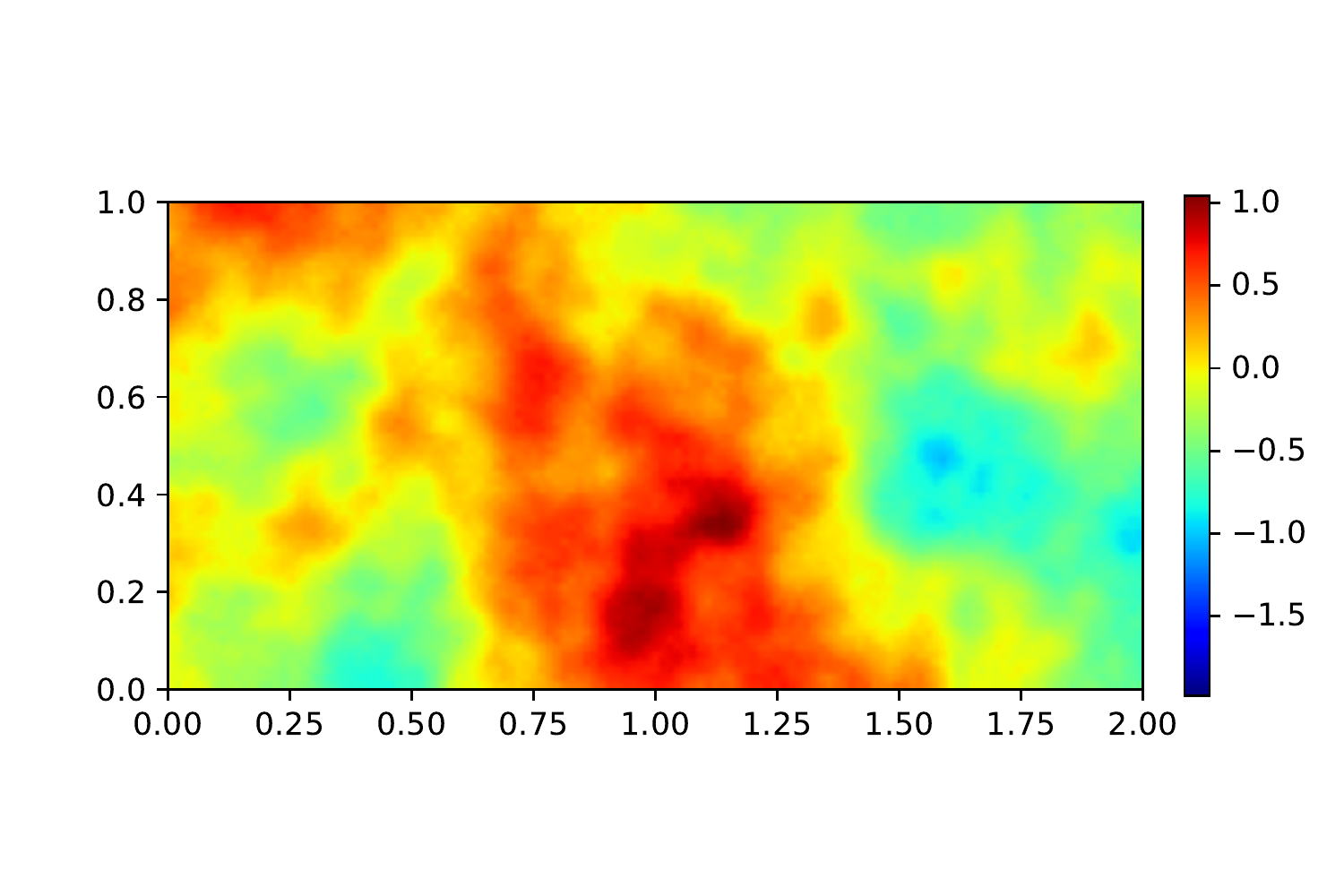}
    & 
    \includegraphics[width=0.50\textwidth]{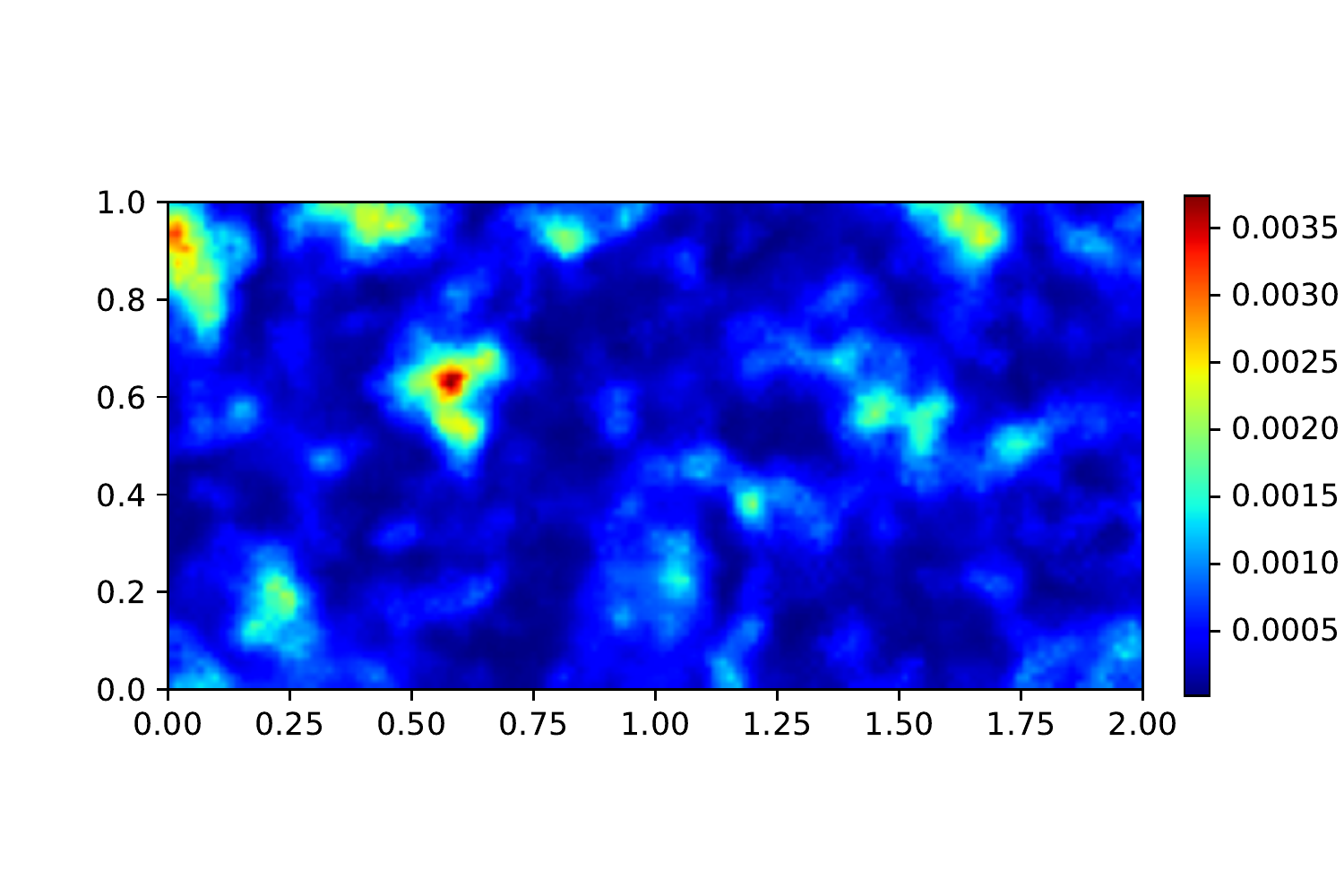}\\
    (e) Mean $\widehat{\bbE}(\xg)$, G-VAE-MCMC. & 
    (f) Variance $\widehat{\bbV}(\xg)$, G-VAE-MCMC.\\
\end{tabular}}
\caption{Estimated mean and variance fields (high-dimensional case).}
\label{fig:compare_high}
\end{figure}

Comparisons of inversion results of the stitched field, the DD-VAE-MCMC method, and the G-VAE-MCMC method for the low and high-dimensional cases are shown in Figure \ref{fig:compare_low} and Figure \ref{fig:compare_high}.
Figure \ref{fig:compare_low}(a),  Figure \ref{fig:compare_low}(c) and Figure \ref{fig:compare_low}(e) compare the estimated mean fields for the low-dimensional case, and Figure \ref{fig:compare_high}(a),  Figure \ref{fig:compare_high}(c) and Figure \ref{fig:compare_high}(e) compare the estimated mean fields for the high-dimensional case. 
It can be seen that the stitched field has clear seams on interfaces $\partial_1\D_2$ and $\partial_2 \D_1$, while the blended image blends the information of two sides and can give visibly intact posterior mean field.
Besides, G-VAE-MCMC fails to give reasonable inversions in both cases.
Also, from Figure \ref{fig:compare_low}(b),  Figure \ref{fig:compare_low}(d) and Figure \ref{fig:compare_low}(f) and Figure \ref{fig:compare_high}(b),  Figure \ref{fig:compare_high}(d) and Figure \ref{fig:compare_high}(f)
 our DD-VAE-MCMC typically gives results with 
smaller
variances.
Heuristically, the Poisson image blending technique, by considering the gradient information, implicitly takes the information of two subdomains into account to give better performance over the global domain.
\begin{table}[!htp]
	\caption{MCMC settings and acceptance rates (low-dimensional).}
	\centering
	\begin{tabular}{cccccc}
		\hline
         Method  &Domain&    Step size&Chain length& Acceptance rate \\
        \hline
        \multirow{2}{*}{DD-VAE-MCMC} &$\D_1$ &  0.04  & 4000& 15.00\%\\
        &$\D_2$  & 0.03 & 4000&  15.90\%\\
                 G-VAE-MCMC & $\D$  & 0.03& 1000& 18.00\%\\
        \hline
	\end{tabular}
	\label{tab:mcmc_low}
\end{table}

\begin{table}[!htp]
	\caption{MCMC settings and acceptance rates (high-dimensional).}
	\centering
	\begin{tabular}{cccccc}
		\hline
         Method  &Domain    &  Step size& Chain length &Acceptance rate \\
        \hline
        \multirow{2}{*}{DD-VAE-MCMC} &$\D_1$ & 0.02  & 4000& 24.47\%\\
        &$\D_2$   & 0.02 & 4000 &  23.38\%\\
        G-VAE-MCMC & $\D$  & 0.03& 1000& 21.90\%\\
        \hline
	\end{tabular}
	\label{tab:mcmc_high}
\end{table}

\begin{table}[!htp]
	\caption{Errors in mean estimate for the two scenarios.}
	\centering
	\begin{tabular}{ccccccc}
		\hline
         Method  &  Low-dimensional case & High-dimensional case \\
        \hline
        DD-VAE-MCMC&  $1.3818\times 10^{-1}$ & $6.2113\times 10^{-1}$ \\
        Stitched field &   $1.4518 \times 10^{-1}$ & $6.6281 \times 10^{-1}$\\
        G-VAE-MCMC &  $3.4009\times 10^{-1}$ & $1.2002$\\
        \hline
	\end{tabular}
	\label{tab:compare_err}
\end{table}

\section{Conclusion}
\label{sec:conclusion}
The principle of divide and conquer is one of the most fundamental concepts for solving high-dimensional Bayesian problems with PDE-involved forward models.
The main difficulties in the Bayesian inverse problems include the curse of dimensionality, implicit prior modeling, and expensive computational costs.
With a particular interest in the case where the prior information is not available in a closed form but only in terms of historical data, this paper proposes a domain-decomposed variational auto-encoder Markov chain Monte Carlo (DD-VAE-MCMC) method. 
VAEs can be utilized to learn the unknown prior distribution with historical data.
Through partitioning the global domain into several subdomains, we propose a DD-VAE framework to represent the local prior distributions in a lower dimensional latent space. 
A GP model with active learning addresses the interface conditions.
The original global problem is then decomposed as a sequence of local problems where inversions can be conducted in lower-dimension parameter spaces and smaller physical subdomains.
After collecting local posterior samples, the Poisson image blending technique is utilized to reconstruct posterior global samples.
Numerical results show that the generative property based on decomposed subdomains is greatly improved, and the efficiency of our proposed method is validated. 

\bigskip
\textbf{Acknowledgments:}
This work is supported by the National Natural Science Foundation of China (No. 12071291), the Science and
Technology Commission of Shanghai Municipality (No. 20JC1414300) and the Natural Science Foundation of Shanghai (No. 20ZR1436200).
\bibliography{xu}

\begin{thebibliography}{10}
\expandafter\ifx\csname url\endcsname\relax
  \def\url#1{\texttt{#1}}\fi
\expandafter\ifx\csname urlprefix\endcsname\relax\def\urlprefix{URL }\fi
\expandafter\ifx\csname href\endcsname\relax
  \def\href#1#2{#2} \def\path#1{#1}\fi

\bibitem{kaipio2006statistical}
J.~Kaipio, E.~Somersalo, Statistical and Computational Inverse Problems, Vol.
  160, Springer Science \& Business Media, 2006.

\bibitem{martin2012stochastic}
J.~Martin, L.~C. Wilcox, C.~Burstedde, O.~Ghattas, A stochastic {Newton MCMC}
  method for large-scale statistical inverse problems with application to
  seismic inversion, SIAM Journal on Scientific Computing 34~(3) (2012)
  A1460--A1487.

\bibitem{wang2005hierachical}
J.~Wang, N.~Zabaras, Hierarchical {Bayesian} models for inverse problems in
  heat conduction, Inverse Problems 21~(1) (2004) 183.

\bibitem{yeh1986review}
W.~W.-G. Yeh, Review of parameter identification procedures in groundwater
  hydrology: The inverse problem, Water resources research 22~(2) (1986)
  95--108.

\bibitem{tarantola2005inverse}
A.~Tarantola, Inverse problem theory and methods for model parameter
  estimation, SIAM, 2005.

\bibitem{stuart2010inverse}
A.~M. Stuart, Inverse problems: a {Bayesian} perspective, Acta numerica 19
  (2010) 451--559.

\bibitem{metropolis1953equation}
N.~Metropolis, A.~W. Rosenbluth, M.~N. Rosenbluth, A.~H. Teller, E.~Teller,
  Equation of state calculations by fast computing machines, The journal of
  chemical physics 21~(6) (1953) 1087--1092.

\bibitem{hastings1970monte}
W.~K. Hastings, {Monte Carlo} sampling methods using {Markov} chains and their
  applications, Biometrika 57~(1) (1970) 97 -- 109.

\bibitem{robert2013monte}
C.~Robert, G.~Casella, {Monte Carlo} statistical methods, Springer Science \&
  Business Media, 2013.

\bibitem{salakhutdinov2015learning}
R.~Salakhutdinov, Learning deep generative models, Annual Review of Statistics
  and Its Application 2 (2015) 361--385.

\bibitem{laloy2017inversion}
E.~Laloy, R.~H{\'e}rault, J.~Lee, D.~Jacques, N.~Linde, Inversion using a new
  low-dimensional representation of complex binary geological media based on a
  deep neural network, Advances in water resources 110 (2017) 387--405.

\bibitem{kingma2013auto}
D.~P. Kingma, M.~Welling, Auto-encoding variational {Bayes}, arXiv preprint
  arXiv:1312.6114.

\bibitem{xia2021bayesian}
Y.~Xia, N.~Zabaras, Bayesian multiscale deep generative model for the solution
  of high-dimensional inverse problems, Journal of Computational Physics 455
  (2022) 111008.

\bibitem{tewari2022subsurface}
A.~Tewari, B.~Wheelock, J.~Clark, D.~Foster, M.~Li, Y.~Marzouk, Subsurface
  uncertainty quantification with deep geologic priors: A variational
  {Bayesian} framework, in: Second International Meeting for Applied Geoscience
  \& Energy, Society of Exploration Geophysicists and American Association of
  Petroleum Geologists, 2022, pp. 1745--1749.

\bibitem{lopez2021deep}
J.~Lopez-Alvis, E.~Laloy, F.~Nguyen, T.~Hermans, Deep generative models in
  inversion: The impact of the generator's nonlinearity and development of a
  new approach based on a variational autoencoder, Computers \& Geosciences 152
  (2021) 104762.

\bibitem{goodfellow2020generative}
I.~Goodfellow, J.~Pouget-Abadie, M.~Mirza, B.~Xu, D.~Warde-Farley, S.~Ozair,
  A.~Courville, Y.~Bengio, Generative adversarial networks, Communications of
  the ACM 63~(11) (2020) 139--144.

\bibitem{elman14finite}
H.~Elman, D.~Silvester, A.~Wathen, {Finite Elements} and {Fast Iterative
  Solvers}: with {Applications} in {Incompressible Fluid Dynamics}, Oxford
  University Press (UK), 2014.

\bibitem{beskos2008mcmc}
A.~Beskos, G.~Roberts, A.~Stuart, J.~Voss, {MCMC} methods for diffusion
  bridges, Stochastics and Dynamics 8~(03) (2008) 319--350.

\bibitem{cotter2013mcmc}
S.~L. Cotter, G.~O. Roberts, A.~M. Stuart, D.~White, {MCMC} methods for
  functions: modifying old algorithms to make them faster, Statistical Science
  28~(3) (2013) 424--446.

\bibitem{dinh2016density}
L.~Dinh, J.~Sohl-Dickstein, S.~Bengio, Density estimation using {Real NVP},
  arXiv preprint arXiv:1605.08803.

\bibitem{ho2020denoising}
J.~Ho, A.~Jain, P.~Abbeel, Denoising diffusion probabilistic models, Advances
  in Neural Information Processing Systems 33 (2020) 6840--6851.

\bibitem{marzouk2007stochastic}
Y.~M. Marzouk, H.~N. Najm, L.~A. Rahn, Stochastic spectral methods for
  efficient {Bayesian} solution of inverse problems, Journal of Computational
  Physics 224~(2) (2007) 560--586.

\bibitem{li2014adaptive}
J.~Li, Y.~M. Marzouk, Adaptive construction of surrogates for the {Bayesian}
  solution of inverse problems, SIAM Journal on Scientific Computing 36~(3)
  (2014) A1163--A1186.

\bibitem{liao2019adaptive}
Q.~Liao, J.~Li, An adaptive reduced basis {ANOVA} method for high-dimensional
  {Bayesian} inverse problems, Journal of Computational Physics 396 (2019)
  364--380.

\bibitem{spanos1989stochastic}
P.~D. Spanos, R.~Ghanem, Stochastic finite element expansion for random media,
  Journal of engineering mechanics 115~(5) (1989) 1035--1053.

\bibitem{chen2015local}
Y.~Chen, J.~Jakeman, C.~Gittelson, D.~Xiu, Local polynomial chaos expansion for
  linear differential equations with high dimensional random inputs, SIAM
  Journal on Scientific Computing 37~(1) (2015) A79--A102.

\bibitem{xu2022domain}
Z.~Xu, Q.~Liao, J.~Li, Domain-decomposed {Bayesian} inversion based on local
  {Karhunen}-{Lo\`{e}ve} expansions, arXiv preprint arXiv: 2211.04026.

\bibitem{liao2015domain}
Q.~Liao, K.~Willcox, A domain decomposition approach for uncertainty analysis,
  SIAM Journal on Scientific Computing 37~(1) (2015) A103--A133.

\bibitem{quarteroni1999domain}
A.~M. Quarteroni, A.~Valli, Domain decomposition methods for partial
  differential equations, Oxford University Press, 1999.

\bibitem{jagtap2020extended}
A.~D. Jagtap, G.~E. Karniadakis, Extended physics-informed neural networks
  {(XPINNs)}: {A} generalized space-time domain decomposition based deep
  learning framework for nonlinear partial differential equations,
  Communications in Computational Physics 28~(5) (2020) 2002--2041.

\bibitem{jagtap2020conservative}
A.~D. Jagtap, E.~Kharazmi, G.~E. Karniadakis, Conservative physics-informed
  neural networks on discrete domains for conservation laws: {Applications} to
  forward and inverse problems, Computer Methods in Applied Mechanics and
  Engineering 365 (2020) 113028.

\bibitem{li2019d3m}
K.~Li, K.~Tang, T.~Wu, Q.~Liao, {D3M}: {A} deep domain decomposition method for
  partial differential equations, IEEE Access 8 (2019) 5283--5294.

\bibitem{li2023deep}
S.~Li, Y.~Xia, Y.~Liu, Q.~Liao, A deep domain decomposition method based on
  {Fourier} features, Journal of Computational and Applied Mathematics 423
  (2023) 114963.

\bibitem{rasmussen2003gaussian}
C.~E. Rasmussen, Gaussian processes in machine learning, in: Summer school on
  machine learning, Springer, 2003, pp. 63--71.

\bibitem{perez2003poisson}
P.~P{\'e}rez, M.~Gangnet, A.~Blake, Poisson image editing, in: ACM SIGGRAPH
  2003 Papers, 2003, pp. 313--318.

\bibitem{alnaes2015fenics}
M.~Aln{\ae}s, J.~Blechta, J.~Hake, A.~Johansson, B.~Kehlet, A.~Logg,
  C.~Richardson, J.~Ring, M.~E. Rognes, G.~N. Wells, The fenics project version
  1.5, Archive of Numerical Software 3~(100).

\bibitem{heusel2017gans}
M.~Heusel, H.~Ramsauer, T.~Unterthiner, B.~Nessler, S.~Hochreiter, {GANs}
  trained by a two time-scale update rule converge to a local nash equilibrium,
  Advances in neural information processing systems 30.

\bibitem{gpy2014}
{GPy}, {GPy}: A gaussian process framework in python,
  \url{http://github.com/SheffieldML/GPy} (since 2012).

\end{thebibliography}
\begin{appendices}
\section{The neural network architecture for the encoder and decoder networks of VAEs.}
\label{appendix:vae_details}
\begin{figure}[!htp]
    \centerline{
    \begin{tabular}{cc}
    \includegraphics[width=0.50\textwidth]{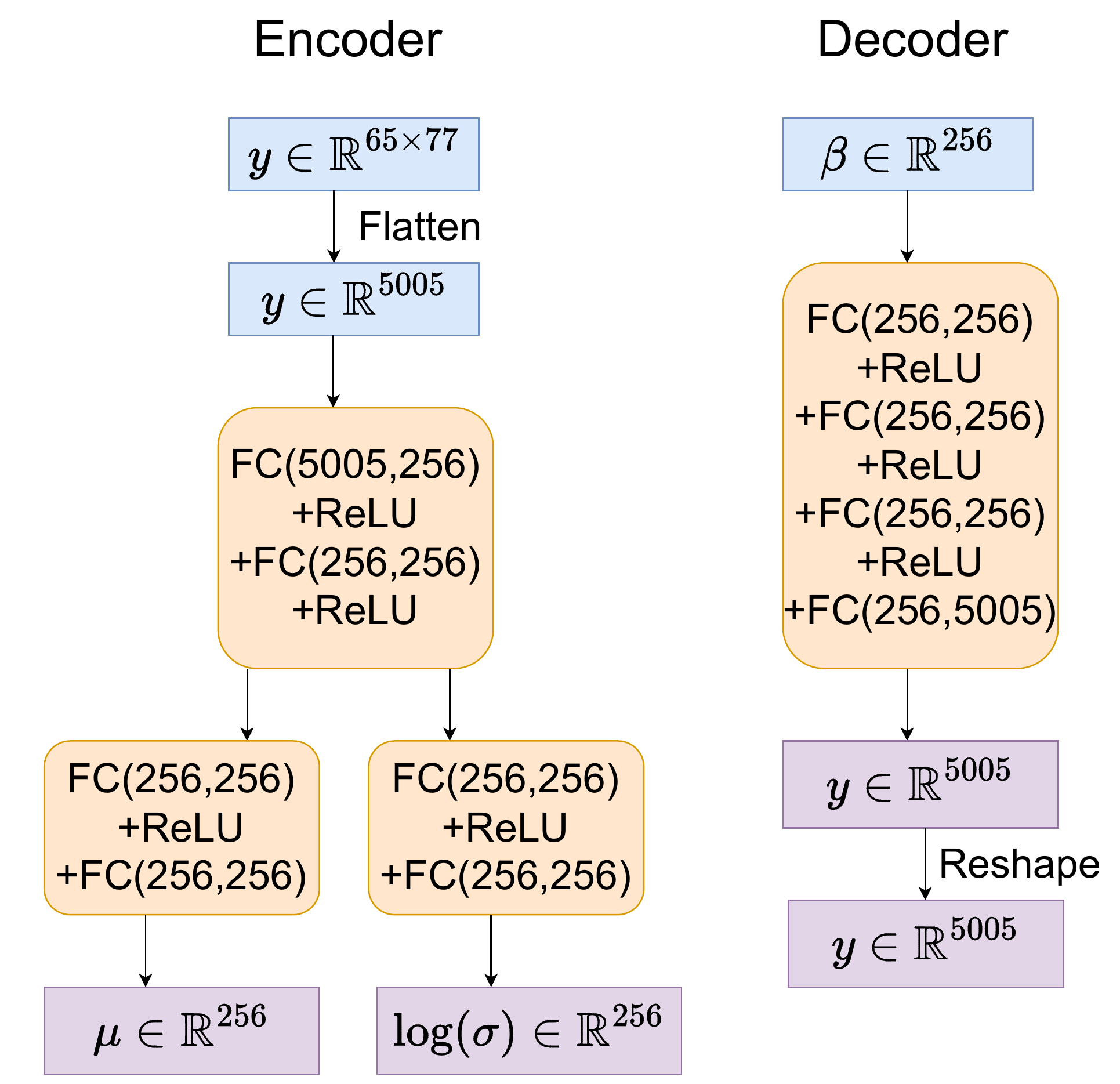}
    & 
    \includegraphics[width=0.50\textwidth]{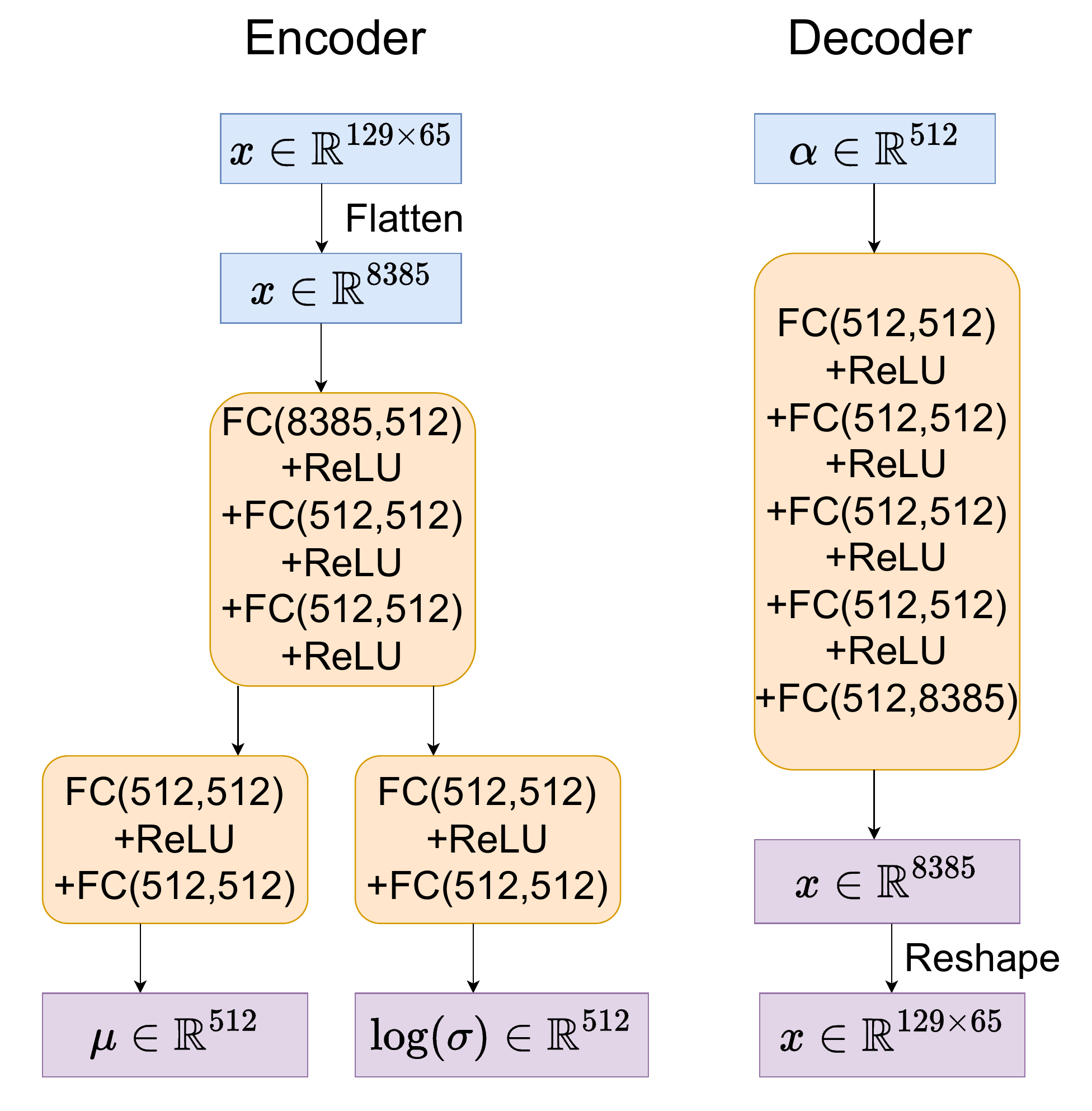}\\
     (a) The network structure A & 
     (b) The network structure B\\
    \end{tabular}}
    \caption{VAE structures (type A and type B) for prior learning.}
    \label{fig:net_struc}
\end{figure}

In this section, we provide details about the network settings we use. 
In this work, we use fully-connected neural networks (FCNN) for the encoder and decoder models in the VAE.  
The network structures we use in this work are depicted in Figure \ref{fig:net_struc}, and the associated hyperparameters used in the training process are listed in Table \ref{tab:hyper}.

\begin{table}[!htp]
	\caption{Neural network settings for VAEs}
	\centering
	\begin{tabular}{c c c}
        \hline 
                       & G-VAE &  DD-VAE \\
                       \hline 
        Architecture & Type B & Type A \\
        Size of training data set & 10000 & 20000 \\
        Epochs &  100 & 100 \\
        Learning rate  &0.001& 0.0001\\
        Batch size &32 & 64\\
        Optimizer &Adam &  Adam\\
        \makecell[c]{Optimizer parameters\\ ($\beta_1, \beta_2$)}  & 0.5, 0.999& 0.5, 0.999  \\
        \hline
	\end{tabular}
	\label{tab:hyper}
\end{table}

\end{appendices}
\end{document}